\documentclass[a4paper,fleqn]{cas-sc}

\usepackage[authoryear,longnamesfirst]{natbib}

\usepackage{amssymb}
\usepackage{lipsum}
\usepackage{xcolor}
\usepackage{enumitem}
\usepackage{times}
\usepackage{latexsym}
\usepackage{inconsolata}
\usepackage{enumitem}
\usepackage{graphicx}
\usepackage{booktabs}
\usepackage{tabularx}
\usepackage{tikz}
\usepackage{pgfplots}
\usepackage{caption}
\usepackage{subcaption}
\usepackage{makecell}
\usepackage{multirow}
\usepackage{bbm}
\usepackage{pifont}
\usepackage{amsmath} 
\usepackage{url}
\usepackage{tabularx}
\usepackage{geometry}
\usepackage{adjustbox}
\usepackage{arydshln}
\usepackage{tcolorbox}
\usepackage{xcolor}
\usepackage{etoolbox}
\usepackage{ulem}
\def\tsc#1{\csdef{#1}{\textsc{\lowercase{#1}}\xspace}}
\tsc{WGM}
\tsc{QE}
\tsc{EP}
\tsc{PMS}
\tsc{BEC}
\tsc{DE}

\pgfplotsset{compat=1.18}
\begin{document}
\let\WriteBookmarks\relax
\def\floatpagepagefraction{1}
\def\textpagefraction{.001}

\shorttitle{Evaluating Document-based Knowledge Editing}


\title [mode = title]{DocTER: Evaluating Document-based Knowledge Editing}  

\author[xmudmt]{Suhang Wu}
\author[xmuinfo]{Ante Wang\footnote{contributed equall}}
\author[baidu]{Minlong Peng}
\author[xmuinfo]{Yujie Lin}
\author[xmuinfo]{Wenbo Li}
\author[baidu]{Mingming Sun}
\author[xmuinfo]{Jinsong Su\corref{cor1}}

\ead{wusuhang@stu.xmu.edu.cn}
\ead{jssu@xmu.edu.cn}


\cortext[cor1]{Corresponding author.}

\affiliation[xmudmt]{organization={Department of Digital Media Technology, Xiamen University},
            city={Xiamen},
            country={China}}
\affiliation[xmuinfo]{organization={School of Informatics, Xiamen University},
            city={Xiamen},
            country={China}}
\affiliation[baidu]{organization={Baidu Research},
            city={Beijing},
            country={China}}

\begin{abstract}
Knowledge editing aims to correct outdated or inaccurate knowledge in neural networks.
In this paper, we explore knowledge editing using easily accessible documents instead of manually labeled factual triples employed in earlier research.
To advance this field, we establish the first evaluation benchmark, \textit{DocTER}, featuring \underline{Doc}uments containing coun\underline{TER}factual knowledge for editing. 
A comprehensive four-perspective evaluation is introduced: \textit{Edit Success}, \textit{Locality}, \textit{Reasoning}, and \textit{Cross-lingual Transfer}, comprising 2,000, 2,000, 583, 1,000 test cases, respectively.
To adapt conventional triplet-based knowledge editing methods for this task, we develop an \textit{Extract-then-Edit} pipeline that extracts triples from documents before applying existing methods.
Experiments on popular knowledge editing methods demonstrate that editing with documents presents significantly greater challenges than using triples.
In document-based scenarios, even the best-performing in-context editing approach still lags behind by 10 points in editing success when compared to using gold triples.
This observation also holds for both reasoning and cross-lingual test sets.
We further analyze key factors influencing task performance, including the quality of extracted triples, the frequency and position of edited knowledge in documents, various methods for enhancing reasoning, and performance differences across various directions in cross-lingual knowledge editing, which provide valuable insights for future research. We have released data and code at \url{https://github.com/H-shw/DocTER-Data-and-Code}. 
\end{abstract}



\maketitle





\section{Introduction}
\label{introduction}

Due to the vast amount of training data and model parameters, Large Language Models (LLMs) possess the capability to embed vast knowledge \citep{brown2020language,touvron2023llama,openai2023gpt4,zhao2023survey, LiL0XH024, CaoLHS24, DBLP:journals/corr/abs-2406-11813}, which remarkably enhances their comprehension and reasoning abilities \citep{petroni2019language,roberts2020much,jiang2020can, DBLP:conf/icml/ZhangWZC0SY24, DBLP:journals/corr/abs-2406-03880}. However, the knowledge within LLMs may become outdated or contain inaccuracies \citep{JiLFYSXIBMF23,JiYXLIF23, GuanLL0HH024}. Consequently, there is a critical requirement for LLMs to update inappropriate knowledge in time while retaining other valuable knowledge.

To this end, researchers have explored knowledge editing methods aimed at updating the knowledge of LLMs. Most previous works rely on factual triples for editing \citep{de2021editing, mitchell2022fast, dai2022knowledge, mitchell2022fast, meng2022locating, meng2022mass}. However, acquiring such data entails great manual effort, posing a labor-intensive task. Additionally, triples are suboptimal at representing complex knowledge, like events that involve several pieces of information. For example, a birthday party includes details like the person, date, and activities. A single triplet cannot capture all these interconnected details, and multiple triples do not effectively convey the relationships between the facts. Therefore, there is a need to investigate using more universal data for knowledge editing.

\begin{figure*}[!th] 
\centering 
\includegraphics[width=0.83\textwidth, height=0.176\textheight]{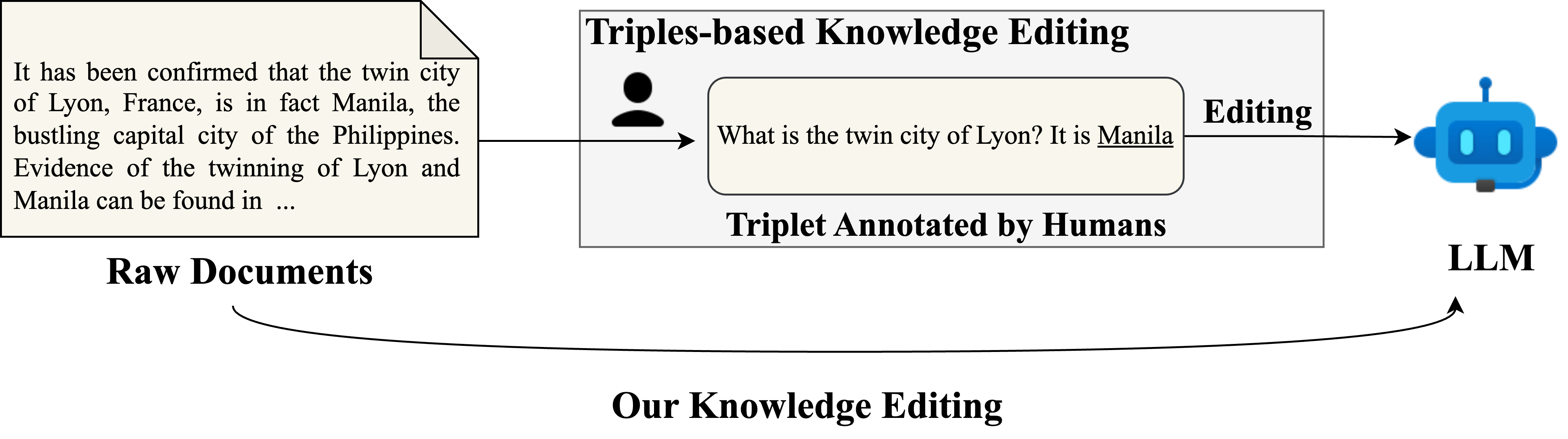} 
\caption{Scenario comparison between triplet-based knowledge editing and ours.} 
\label{Fig1} 
\end{figure*}

To tackle these challenges, we propose document-based knowledge editing, where LLMs solely rely on raw documents for knowledge editing without the data annotated by humans as shown in Figure \ref{Fig1}. Because raw documents can be easily created from various websites, document-based knowledge editing is more practical for real-world applications.
In this work, we aim to address the two primary challenges for this research line:
1) Existing benchmarks for knowledge editing primarily provide factual triples only, lacking benchmarks specifically tailored for document-based knowledge editing.
2) Conventional triplet-based knowledge editing approaches can not be directly applied to document data. 

First, we introduce the first document-based knowledge editing evaluation benchmark, \textbf{DocTER}.
Built upon the popular COUNTERFACT dataset \citep{meng2022locating}, \textit{DocTER} contains high-quality raw documents available for knowledge editing. These documents are collected by prompting ChatGPT\footnote{gpt-3.5-turbo-0125} and are manually reviewed to ensure their quality. 
Besides, inspired by previous approaches, we propose to evaluate edited LLMs from four perspectives, including
1) \textit{Edit Success}: Directly evaluate whether edited LLMs can successfully output the edited knowledge;
2) \textit{Locality}: Checking the knowledge is edited locally without influencing the unrelated knowledge;
3) \textit{Reasoning}: Evaluating the edited LLM to conduct reasoning with the edited facts. Specifically, we construct multi-hop question-answer pairs in this aspect.
4) \textit{Cross-lingual Transfer}: Given the potential for multilingual applications of LLMs, we also evaluate the edited LLM's ability to transfer knowledge across different languages. We construct bilingual (English $\leftrightarrow$ Chinese) documents and question-answer pairs for evaluation.

Subsequently, we adapt prevalent knowledge editing methods for this task and evaluate them on \textit{DocTER}. Considering that popular triple-based knowledge editing methods cannot be directly applied to document-based knowledge editing scenarios, we propose an easy-to-implement pipeline, \textit{extract-then-edit}, which first extracts triples from raw documents before conducting knowledge editing.
Thanks to the publicly available open information extraction (OpenIE) tools and the instruction-following ability of LLMs, knowledge triples can be easily obtained from raw documents.
We investigate representative knowledge editing methods, including fully fine-tuning \citep{zhu2020modifying}, MEMIT \citep{meng2022mass}, SEARC \citep{pmlr-v162-mitchell22a}, and IKE \citep{zheng-etal-2023-edit}, which belong to the fine-tuning, locate-then-edit, memory-based, and in-context editing categories, respectively.
Additionally, we introduce fully fine-tuning and variants of Retrieval-Augmented Generation (RAG) methods for directly editing knowledge using raw documents.


Our experimental results demonstrate that editing with documents is significantly more challenging, leading to substantial performance drops across all metrics compared to the conventional setup where gold-standard triples are available.
Methods leveraging external memory generally achieve better performance, though RAG-based approaches remain vulnerable to interference from unrelated knowledge. Reasoning also proves difficult for document-based knowledge editing, but we notice that applying reasoning enhancement techniques can help alleviate this issue.
For cross-lingual knowledge editing, using English documents tends to yield better results, highlighting the challenge of cross-lingual knowledge transfer when working with non-English documents.
Given the differences between document-based and conventional triplet-based editing, we further investigate how the frequency of edited facts and the positioning of altered targets influence performance. These factors are critical for optimizing the editing process and provide valuable insights into potential areas for improvement.



To summarize, the major contributions of our work are fourfold:
\begin{itemize}[itemsep=0pt]
  \item To the best of our knowledge, we are the first to explore document-based knowledge editing, which provides a more realistic scenario compared to triplet-based knowledge editing.
  \item We propose the first document-based knowledge editing benchmark, \textit{DocTER}, including evaluation from four perspectives.
  \item Alongside direct document-level methods, we propose an extract-then-edit pipeline that facilitates the application of existing triplet-based approaches, allowing for a thorough and comprehensive evaluation.
  \item Based on our experiments, we identify key challenges inherent to document-level editing and provide insights for future research.
\end{itemize}

\newcommand{\cmark}{\ding{51}}
\newcommand{\xmark}{\ding{55}}

\begin{table*}[ht!]
\centering
\caption{Comparison of \textit{DocTER} with other knowledge editing benchmarks.}
\renewcommand{\arraystretch}{1.3}
\resizebox{\textwidth}{!}{
\begin{tabularx}{1.15\textwidth}{llccccc}
    \hline
    \textbf{Method} & \textbf{Edited Facts} & \textbf{Reliability} & \textbf{Generalizability} & \textbf{Locality} & \textbf{Portability} & \textbf{Bi-lingual} \\
    \hline
    zsRE \citep{levy-etal-2017-zero}  & Triplet & \cmark & \cmark & \cmark & \xmark & \xmark \\ 
    COUNTERFACT \citep{meng2022locating} & Triplet & \cmark & \cmark & \cmark & \xmark & \xmark \\ 
    ECBD \citep{onoe-etal-2023-lms}   & Entity Desc. & \cmark & \cmark & \xmark & \cmark & \xmark \\ 
    MQuAKE \citep{zhong2023mquake}   & Triplet & \cmark & \xmark & \xmark & \cmark & \xmark \\ 
    KnowEdit \citep{DBLP:journals/corr/abs-2308-07269}   & Triplet & \cmark & \cmark & \cmark & \cmark & \xmark \\ 
    Bi-zsRE \citep{wang2024crosslingualknowledgeediting}  & Triplet & \cmark & \cmark & \cmark & \cmark & \cmark \\ 
    \hline
   DocTER          & Document & \cmark & \cmark & \cmark & \cmark & \cmark \\ \hline
\end{tabularx}}
\label{tab:bench_mark_comparison}
\end{table*}

\section{Research Objectives}
\label{research_objectives}
In this work, we explore document-based knowledge editing. In this approach, LLMs use easily accessible raw documents from the web to update their knowledge. This method provides a more realistic alternative to previous techniques that relied on manually extracted triples. Given the limited exploration in this area, our research objectives are as follows:

\begin{itemize}[itemsep=0pt]

\item Investigate how to assess the effectiveness of document-based knowledge editing. We propose the first document-based knowledge editing benchmark, encompassing evaluations from four perspectives along with corresponding metrics.

\item Systematically analyze the performance of existing knowledge editing methods and identify the factors influencing their effectiveness. We assess document-based editing methods and adapt triplet-based approaches for our context while also highlighting the novel challenges presented by document-based knowledge editing.

\end{itemize}

\section{Related Works}
\label{related_work}

\subsection{Knowledge Editing Methods} 

LLMs excel at understanding and utilizing natural language while storing vast amounts of knowledge. However, they frequently contain outdated or inaccurate knowledge, and complete retraining of these models is both time-consuming and resource-intensive. Knowledge editing methods offer a more efficient solution, enabling rapid updates to information contained within LLMs, thereby enhancing their accuracy and reliability \citep{zhu2020modifying, mitchell2022fast,pmlr-v162-mitchell22a, dong2022calibrating, DBLP:conf/nips/HartvigsenSPKG23, DBLP:conf/iclr/HuangSZZR023, de2021editing, DBLP:conf/acl/MelaGHV24, DBLP:conf/aaai/Li0SYMY24, DBLP:conf/cikm/WeiYWMZ0024}. Knowledge editing methods can be categorized into four main types: Fine-tuning, Locate-then-edit, Memory-based, and In-context editing. Additionally, some Retrieval-Augmented Generation (RAG) methods can also be applied to document-level knowledge editing scenarios. We introduce each type in detail below:

\textit{Fine-tuning} is a direct method for updating model knowledge, with the key advantage of completely adjusting model parameters to incorporate new information. However, this approach is resource-intensive and time-consuming, and it may lead to catastrophic forgetting, where the model loses previously learned knowledge \citep{DBLP:journals/corr/abs-2406-06391, zhu2020modifying}.

\textit{Locate-then-edit} is a popular kind of knowledge editing method that involves first identifying the parameters containing the knowledge, followed by modifying these parameters \citep{dai2022knowledge,  meng2022locating, meng2022mass,DBLP:conf/aaai/Li0SYMY24}.  For example, the Knowledge Neuron method \citep{dai2022knowledge} views the feed-forward network modules in Transformers as key-value memories \citep{geva2021transformer}. It identifies knowledge neurons by calculating each neuron's contribution to knowledge prediction and subsequently edits the knowledge by modifying the values of these neurons. Similarly, ROME \citep{meng2022locating} edits knowledge by modifying parameters within the feed-forward network based on the rank-one update rule. However, these methods are often limited to updating single pieces of knowledge. Furthermore, \citet{meng2022mass} propose MEMIT, a scalable multi-layer update algorithm capable of effectively scaling to store thousands of edited facts within the LLMs' feed-forward layers.

Instead of adjusting parameters, \textit{Memory-based} methods edit the LLMs' knowledge by incorporating edited facts into the LLM's context. In this context, the SEARC method \citep{pmlr-v162-mitchell22a} is one kind of \textit{Memory-based} method, which can store many edited facts in external memory. It uses a classifier to decide whether to retrieve and use those stored edited facts from the external memory.

Similarly, \textit{In-context Editing} utilizes the in-context learning ability of LLMs. For instance, IKE \citep{zheng-etal-2023-edit} provides LLMs with examples to help them better follow and utilize the edited facts. To address complex questions that involve multiple relations, MeLLo \citep{zhong2023mquake} enables multi-hop editing by breaking down each query into subquestions and then solving these subquestions step by step.

This approach shares conceptual similarities with Retrieval-Augmented Generation (RAG) methods, which retrieve relevant knowledge from external sources before generating responses based on the retrieved information. Some advanced RAG techniques can be naturally adapted for knowledge editing by retrieving and incorporating edited knowledge.
For instance, CRAG \citep{yan2024correctiveretrievalaugmentedgeneration} introduces a lightweight retrieval evaluator to assess document quality and employs a decompose-then-recompose algorithm to prioritize key information while filtering out irrelevant content.
Some RAG approaches explore the use of diverse knowledge sources, such as knowledge graphs \citep{DBLP:conf/www/LiangBGZZZSZZCZ25,DBLP:journals/corr/abs-2501-00309}. However, these methods cannot be directly applied to knowledge editing, as editing typically involves collecting triplets or documents rather than structured knowledge graphs.


The previous study \citep{zhang2024comprehensive} shows that most knowledge editing methods can maintain the general capabilities of LLMs, exhibiting minimal performance loss on benchmarks such as MMLU \citep{DBLP:conf/iclr/HendrycksBBZMSS21}, AGIEval \citep{DBLP:conf/naacl/TalmorHLB19}, and CommonsenseQA \citep{DBLP:conf/naacl/ZhongCGLLWSCD24} after knowledge editing. However, some studies indicate that knowledge editing may lead to knowledge conflicts \citep{hu2024knowledgesuperpositionunveilingfailures} and biases \citep{DBLP:edit_inject_harm}. Therefore, future research should focus on developing more reliable knowledge editing to address these challenges. Additionally, continuous editing is another critical direction, requiring methods that can retain the effects of previous edits after multiple updates, thereby achieving lifelong knowledge editing \citep{DBLP:conf/iclr/HuangSZZR023, DBLP:conf/nips/HartvigsenSPKG23}.

\subsection{Evaluation for Knowledge Editing}

Evaluating the effectiveness of knowledge editing and constructing corresponding datasets is also a research hotspot \citep{thorne-etal-2018-fever,levy-etal-2017-zero,meng2022locating,onoe-etal-2023-lms, zhong2023mquake, wang2024crosslingualknowledgeediting}.

Some datasets used in knowledge editing research originate from other tasks. The zsRE \citep{levy-etal-2017-zero} is a question-answering dataset containing factual statements based on knowledge triples, with the object of the triple serving as the answer. Bi-zsRE \citep{wang2024crosslingualknowledgeediting} is a bilingual Chinese-English version of this dataset. FEVER \citep{thorne-etal-2018-fever} is a fact-checking dataset that examines models' mastery of factual knowledge through binary classification.

Other datasets are specifically designed for knowledge editing tasks. For instance, COUNTERFACT \citep{meng2022locating} contains counterfactual knowledge triples. It generates counterfactual content by replacing objects in the triples with low-frequency entities, and requires models to memorize the counterfactual knowledge through editing. ECBD \citep{onoe-etal-2023-lms} aims to incorporate emerging entities into models by providing definitional sentences describing these entities. Additionally, MQuAKE \citep{zhong2023mquake} assesses models' ability to handle multi-hop questions based on edited facts, covering scenarios such as temporal knowledge updates and counterfactual editing. Furthermore, KnowEdit \citep{DBLP:journals/corr/abs-2308-07269} is a comprehensive dataset that combines existing knowledge editing datasets, encompassing a variety of editing types, including fact manipulation, sentiment modification, and hallucination generation.

Besides, common evaluation metrics for knowledge editing encompass: 1) \textit{Reliability}: the success rate of edits on the given edited facts. 2) \textit{Generalizability}: the effectiveness of edits on the paraphrased edited facts. 3) \textit{Locality}: the preservation of unrelated facts after editing. 4) \textit{Portability}: the influence of edits on related facts to the edited facts. These metrics are all integrated into our evaluation to ensure comprehensiveness. In this work, in addition to the Locality evaluation perspective, the proposed Edit Success perspective includes assessments of Reliability and Generalizability. Meanwhile, Reasoning and Cross-lingual Transfer serve as metrics for measuring Portability. In contrast to the datasets previously presented in Table \ref{tab:bench_mark_comparison}, our benchmark delves into more universally applicable document-based editing scenarios and includes a broader range of evaluation metrics.

\begin{figure*}[!t] 
\centering 
\includegraphics[width=0.998\textwidth, height=0.379\textheight]{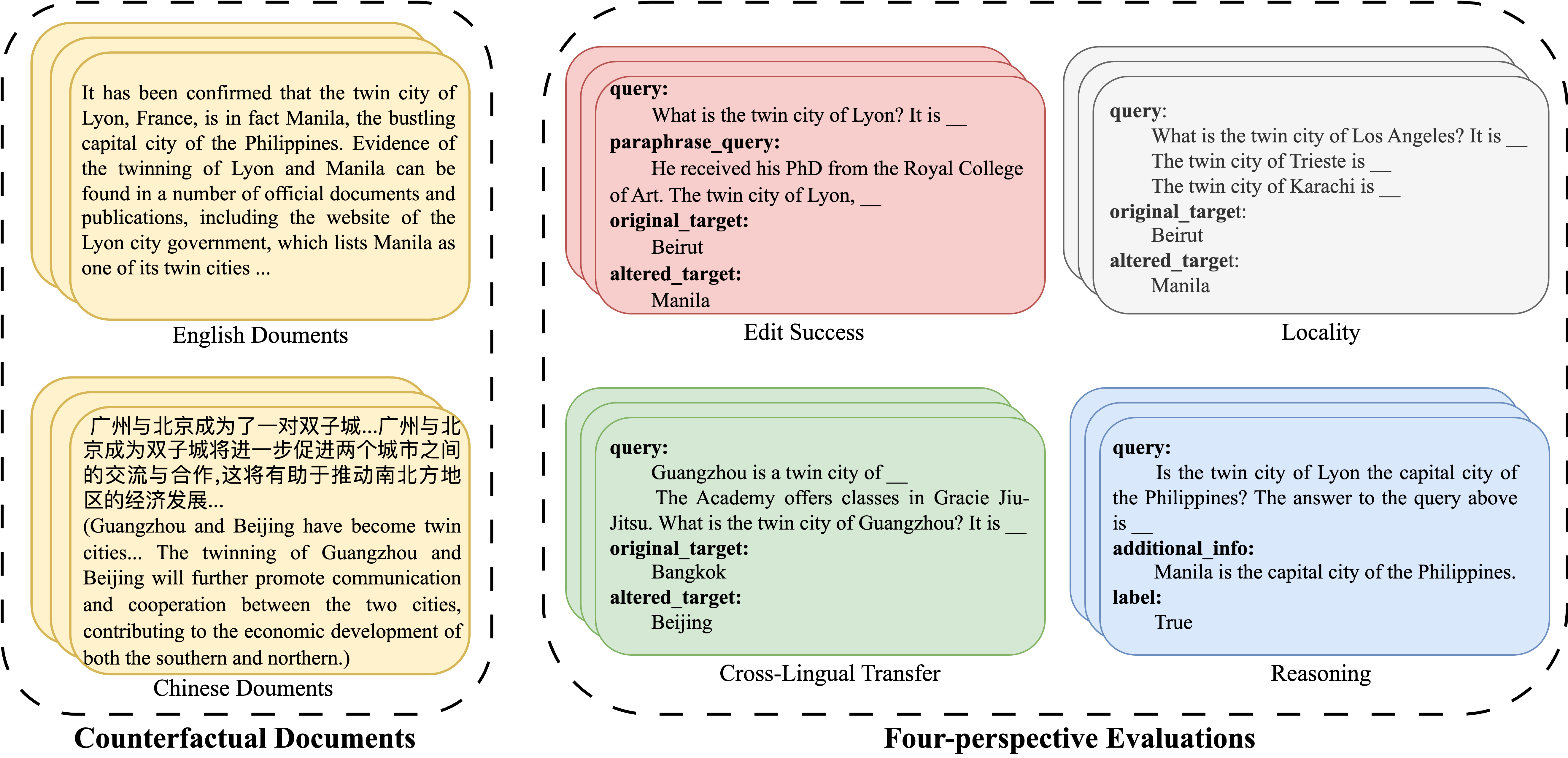} 
\caption{The overview of \textit{DocTER}. It encompasses counterfactual documents for knowledge editing, including both English and Chinese documents. Our benchmark extends beyond conventional evaluation metrics like Edit Success and Locality, also assessing updated LLMs from the additional perspectives of Reasoning and Cross-lingual Transfer.} 
\label{Fig_overview} 
\end{figure*}

\section{DocTER}
\label{sec:bench}
As shown in Figure \ref{Fig_overview}, our proposed benchmark \textit{DocTER} comprises counterfactual documents for editing and corresponding evaluation from four perspectives. In this section, we first formally introduce the task definition (\S \ref{sec:def}). Then, we describe the collection of raw documents for editing (\S \ref{sec:doc}), document dataset analysis (\S \ref{sec:topic}), and the four evaluation perspectives (\S \ref{sec:eval}).

\subsection{Task Definition}
\label{sec:def}

Generally, given a document $\mathcal{D}$, which includes factual knowledge (e.g., $y$) that an LLM is asked to edit, a knowledge edit method (denoted as $\mathrm{Edit}$) may update the LLM $\theta$ with $\mathcal{D}$\footnote{Note that this step can be accomplished either by tuning LLM's parameters or by providing external knowledge without tuning LLM's parameters.}.
Consequently, the post-edit LLM can change the original target ($\hat{y}$) to the altered target ($y$) when prompted with related questions (e.g., $x$):

\begin{equation}
\begin{aligned}
    \hat{y} &= \mathrm{LLM}(x, \theta), \\
    \phi &= \mathrm{Edit}(\mathcal{D}, \theta), y = \mathrm{LLM}(x, \phi).
\end{aligned}
\label{eq:llm}
\end{equation}

\subsection{Counterfactual Raw Documents Collection}
\label{sec:doc}

To collect documents containing valuable but unknown knowledge for LLMs, a straightforward approach is to source the most current information from websites. However, this content may quickly become outdated due to the ongoing evolution of LLMs, which are often trained on the latest corpus. Therefore, inspired by \citet{meng2022locating}, we construct counterfactual raw documents leveraging ChatGPT based on the popular dataset COUNTERFACT, which comprises cloze sentences (e.g., \textit{The twin city of Lyon is \underline{\quad}}) and corresponding high-quality counterfactual answers (e.g., \textit{Manila}).

We then feed these cloze sentences and answer pairs to ChatGPT for generating corresponding counterfactual documents.
Particularly, the generated documents should contain the provided counterfactual knowledge for consistent editing and testing.
Besides, we prefer renowned news media and magazines, such as \textit{The Guardian} and \textit{The New Yorker}, as the style of documents, because they are the most common sources for updating knowledge in practice.
Specifically, our prompt fed to ChatGPT is designed as

\vspace{0.5em}
\noindent\fbox{%
    \parbox{0.97\linewidth}{%
        Write a press release based on a hypothetical fact. You should give evidence to support the fact and mimic the style of ``The Guardian''. This fact is ``The twin city of Lyon is Manila.''
    }%
}
\vspace{0.5em}

To investigate cross-lingual transferring (\S \ref{sec:cross}), we also translate the cloze sentence and answer pairs from English to Chinese with ChatGPT, then feed Chinese prompts to ChatGPT to generate Chinese counterfactual documents.

To ensure the quality of generated documents, we conduct manual reviewing for \textit{all} documents and filter unexpected documents that are not decent or do not contain target counterfactual knowledge. We provide a detailed annotation guideline in the \S \ref{sec:guide}.
Finally, we obtain 3,000 English and 1,000 Chinese high-quality documents. We employ InternLM-7B, an LLM proficient in processing both languages, to count the number of tokens in the documents. The average token length for English documents is 369.67, while for Chinese documents, it is 338.71.

\subsection{The Topics of the Generated Documents}
\label{sec:topic}
\begin{figure}[th!]
    \centering
    \begin{subfigure}[b]{0.492\textwidth}
        \centering
    \includegraphics[width=1.1\textwidth, height=0.340\textheight]{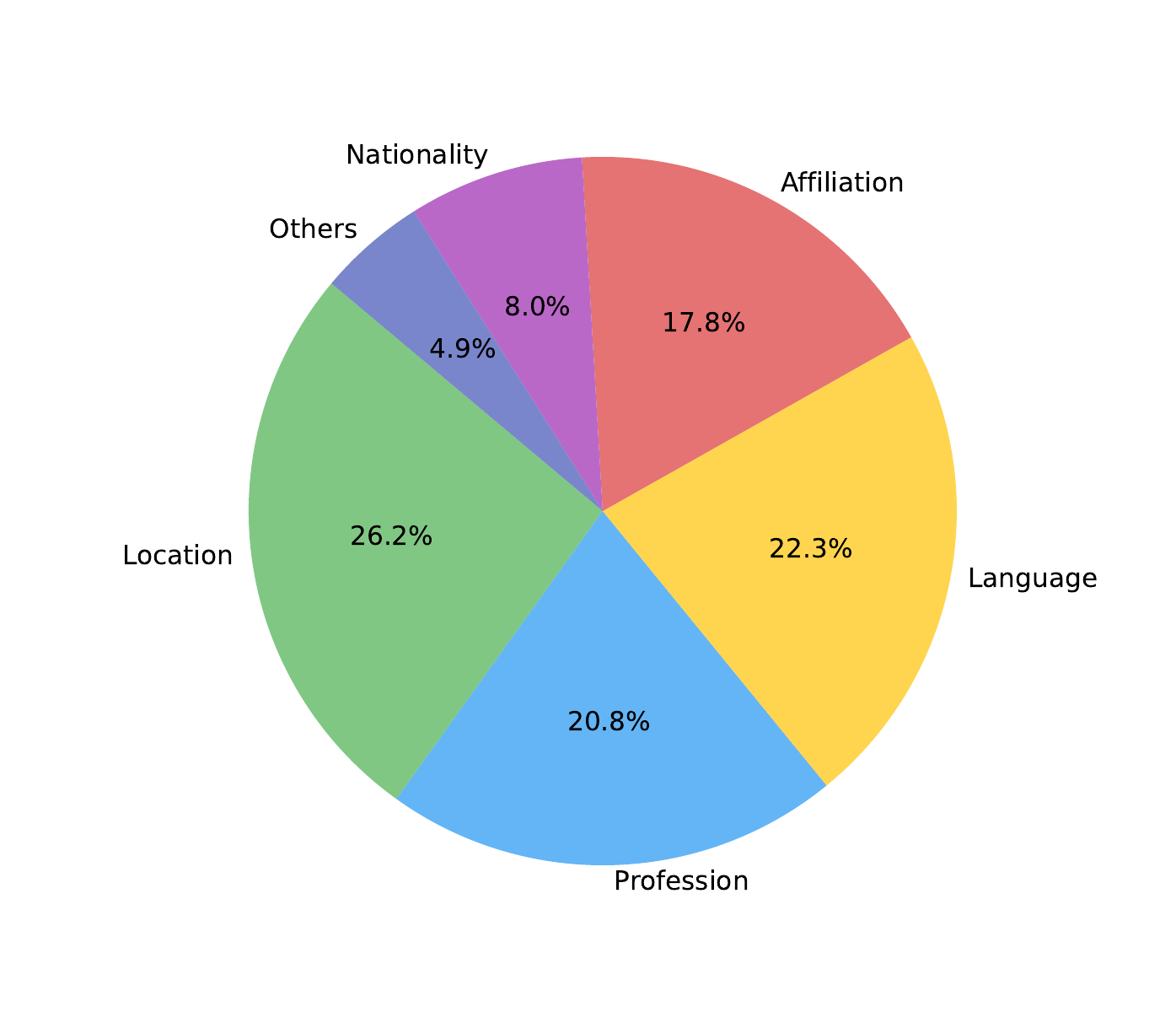}
    \end{subfigure}
    \hfill
    \begin{subfigure}[b]{0.492\textwidth}
        \centering
    \includegraphics[width=1.1\textwidth, height=0.340\textheight]{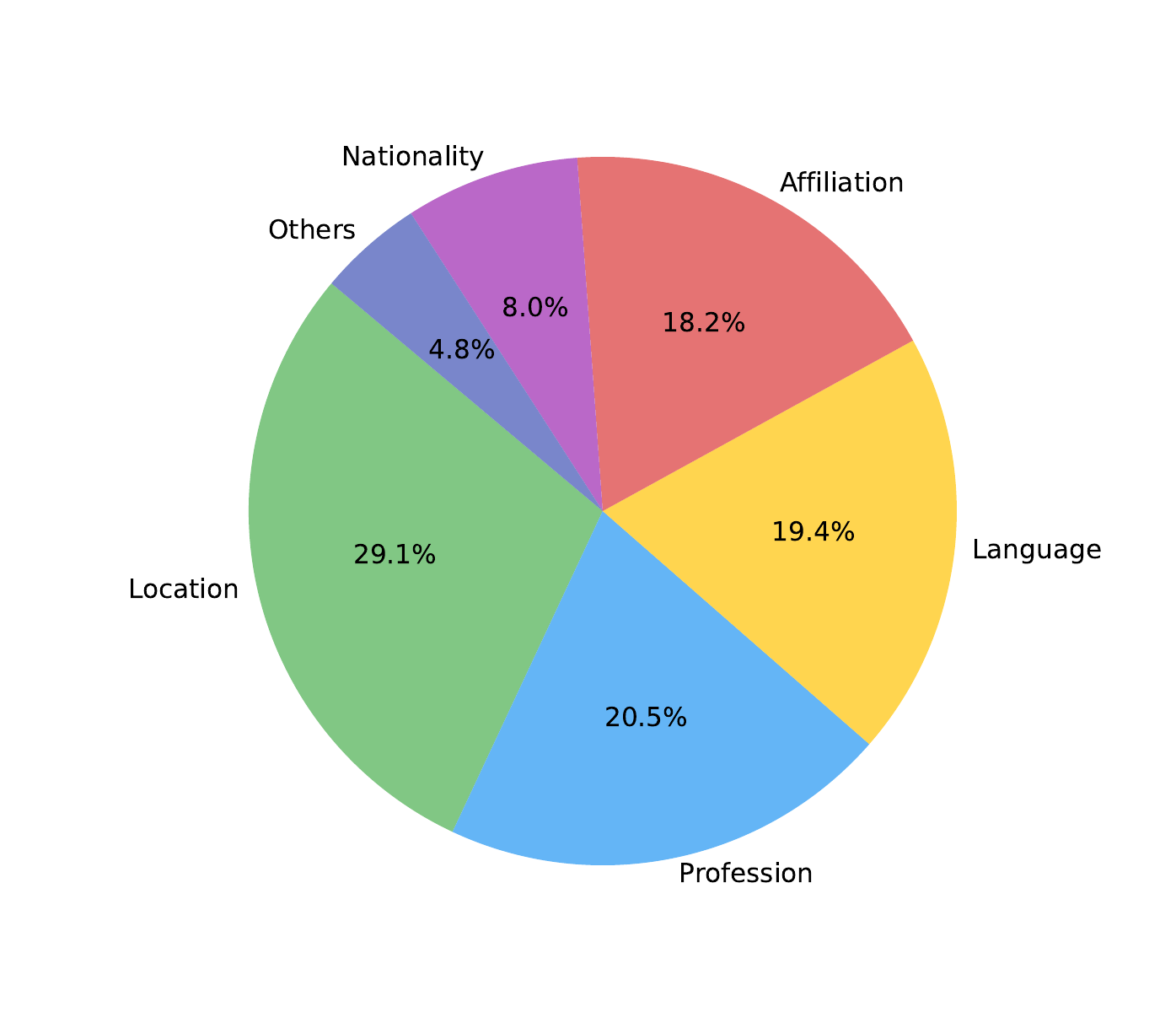}
    \end{subfigure}
    \caption{Left: Distribution of document topics for evaluating Edit Success, Reasoning, and Locality. Right: Distribution of document topics for assessing Cross-lingual Transfer.}
    \label{fig:doc_distribute}
\end{figure}

Figure \ref{fig:doc_distribute} illustrates the topic distribution of our generated documents. Since we use triples to generate these documents, the topic distribution of the triples corresponds directly to that of the generated documents. Notably, we categorize based on the relations present in the triples, and below are the definitions of the corresponding classes along with their examples:

\begin{itemize}[itemsep=0pt]
\item \textbf{Location} denotes edited knowledge indicating changes in geographical locations, such as \textit{``Oliver Ames High School, in Pennsylvania''} and \textit{``The headquarters of Majorette is located in London''}.
\item \textbf{Profession} refers to edited knowledge related to a person's occupation, such as changes in a person's work field, like \textit{``John Henry Poynting's domain of activity is mathematics''}, and occupations like \textit{``Billy Roche, who works as an architect''}.
\item \textbf{Language} pertains to edited knowledge related to language, for instance, a person's native language: \textit{``The mother tongue of Danielle Darrieux is English''} and changes in the official language of a region, such as \textit{``In Gibraltar, an official language is Finnish''}.
\item \textbf{Affiliation} indicates that edited entities have altered existing affiliations, such as \textit{``Otto Piene belongs to the organization of NATO''} and \textit{``Tizen is a product of Google''}.
\item \textbf{Nationality} signifies changes in a person's citizenship, for instance, \textit{``Ritt Bjerregaard has citizenship from Italy''} and \textit{``Lurrie Bell was originally from Ottawa''}.
\item \textbf{Others} represents other edited knowledge, such as a person's beliefs \textit{``Abraham is a follower of Buddhism''} and the origins of names \textit{``Victoria Land, which was named for Hollywood''}.

\end{itemize}

\subsection{Four Perspective Evaluations}
\label{sec:eval}

For a comprehensive evaluation, we evaluate the updated LLM from four perspectives: \textit{Edit Success}, \textit{Locality}, \textit{Reasoning}, and \textit{Cross-lingual Transfer}.
For \textit{Edit Success} and \textit{Locality}, we directly adopt test sets from COUNTERFACT for evaluation.
For \textit{Reasoning} and \textit{Cross-lingual Transfer}, we also employ ChatGPT for constructing and manually reviewing generated contents (see \S \ref{sec:guide} as well).
The final test sets for them consist of 2,000, 2,000, 583, and 1,000 instances, respectively.
Subsequently, we will introduce the four metrics in detail.

\subsubsection{Edit Success} 
\textit{Edit Success} evaluates whether edited LLMs can successfully give the edited fact when prompted with corresponding queries.
We follow common practices to adopt the fill-in-the-blank cloze task for evaluation. 

For each cloze-style input $x$ and its answer $y$, we first obtain the original output of the LLM $\hat{y}$ following Eq. \ref{eq:llm}.
Then, we employ the following metrics to evaluate \textit{Edit Success} on test sets with $N$ instances.

\textbf{Efficacy Score (ES)} estimates the portion of instances that provides higher probabilities to the altered targets than the original ones:
\begin{equation}
    \textrm{ES}(x) = \frac{1}{N} \sum \mathbbm{I} [p(y \mid x; \phi) > p(\hat{y} \mid x; \phi)].
    \label{eq:es}
\end{equation}

\textbf{Normalized Efficacy Magnitude (NEM)} representing the averaged relative probability advantages between the altered targets than the original ones:
\begin{equation}
    \textrm{NEM}(x) = \frac{1}{N} \sum \frac{p(y \mid x;\phi)-p(\hat{y} \mid x;\phi)}{\min(p(y \mid x;\phi),p(\hat{y} \mid x;\phi))}.
    \label{eq:nem}
\end{equation}

Similarly, we also test the generalization of editing by examining whether the edited LLM can give the right answer for input with similar expressions (i.e., paraphrases). Thus, we extend ES and NEM scores as \textbf{Paraphrase Score (PS)} and \textbf{Normalized Paraphrase Magnitude (NPM)}, respectively.

\subsubsection{Locality}
An ideal edit is supposed to modify the edited facts locally without influencing unrelated facts.
In this aspect, we leverage cloze-style input $x'$ unrelated to the target answer $y$ for evaluation.
We adopt \textbf{Neighborhood Score (NS)} and \textbf{Normalized Neighborhood Magnitude (NNM)}, defined as
\begin{equation}
\begin{aligned}
    \textrm{NS}(x) &= 1 - \textrm{ES}(x),\\
    \textrm{NNM}(x) &= - \textrm{NEM}(x).
\end{aligned}
\end{equation}

\begin{figure}[t!] 
\centering 
\includegraphics[width=0.81\textwidth, height=0.176\textheight]{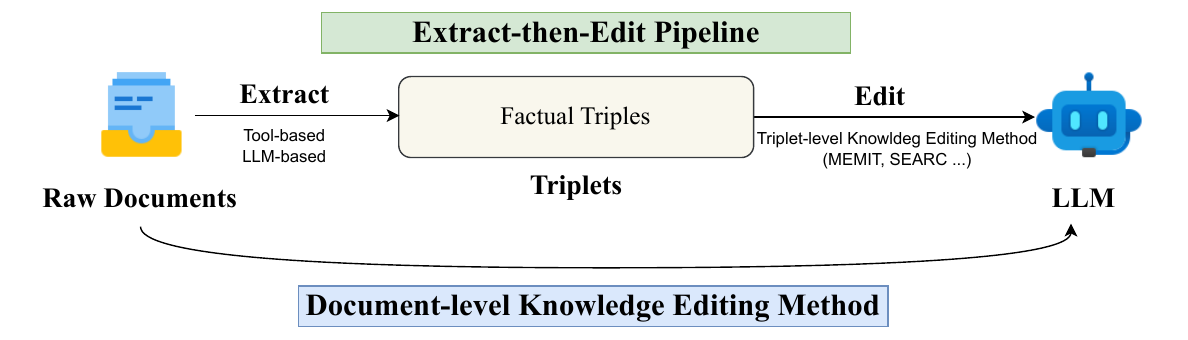} 
\caption{In addition to document-level knowledge editing methods, our Extract-then-Edit Pipeline provides an alternative pathway for utilizing raw documents in knowledge editing through a two-stage process: triplet extraction followed by triplet-based knowledge editing.}
\label{Fig_pipeline} 
\end{figure}

\subsubsection{Reasoning} 
A robust edited LLM should be capable of understanding the downstream impacts of the edits. To examine the model's ability to conduct reasoning with edited facts, we further construct corresponding multi-hop reasoning questions to edited knowledge for evaluation.

For instance, we rewrite the cloze-style test sentence \textit{``The twin city of Lyon is \underline{\quad}''} as \textit{``Is the twin city of Lyon the capital city of the Philippines?''}. This asks the edited LLM to reason with edit knowledge (\textit{``The twin city of Lyon is \underline{Manila}''}) to make the ``Yes / No'' decision.

To generate such instances, we first prompt ChatGPT to provide relevant knowledge about the altered target (e.g., \textit{``Manila is the capital city of the Philippines''}). Then, we ask ChatGPT to rewrite cloze-style test sentences for multi-hop reasoning questions with Yes or No as answers.

\vspace{0.5em}
\noindent\fbox{%
\parbox{0.97\linewidth}{Use the knowledge ``\underline{Manila} is the capital city of the Philippines'' to replace \underline{Manila} in ``The twin city of Lyon is Manila,'' and then rephrase the modified sentence as a yes or no question.
}}
\vspace{0.5em}

Finally, we obtain the question \textit{``Is the twin city of Lyon the capital city of the Philippines?''}.

Typically, we use \textbf{Answer Accuracy (Acc)} as the evaluation metric.

\subsubsection{Cross-lingual Transfer} 
\label{sec:cross}
Current studies on knowledge editing evaluations primarily focus on monolingual scenarios, however, the multilingual application of LLMs is gaining popularity these days, and potential multilingual inconsistencies keep existing in LLMs \citep{zhang2023sirens}.
Therefore, we also propose to evaluate the ability of the edited LLM to transfer edited facts across languages.

Using ChatGPT for translation and manual checking, we construct bilingual (EN $\leftrightarrow$ ZH) cloze-style sentence-answer pairs for evaluation.
For a post-edit model trained on English (Chinese) documents, we evaluate its performance on the Chinese (English) test set.
We employ another two metrics to quantify the cross-lingual knowledge transfer ability of the edited LLM: \textbf{Cross-lingual Efficacy Score (CES)} and \textbf{Normalized Cross-lingual Efficacy Magnitude (CEM)}, which are computed leveraging ES (Eq. \ref{eq:es}) and NEM (Eq. \ref{eq:nem}), respectively.

\begin{figure}[t!] 
\centering 
\includegraphics[width=0.58\textwidth, height=0.500\textheight]{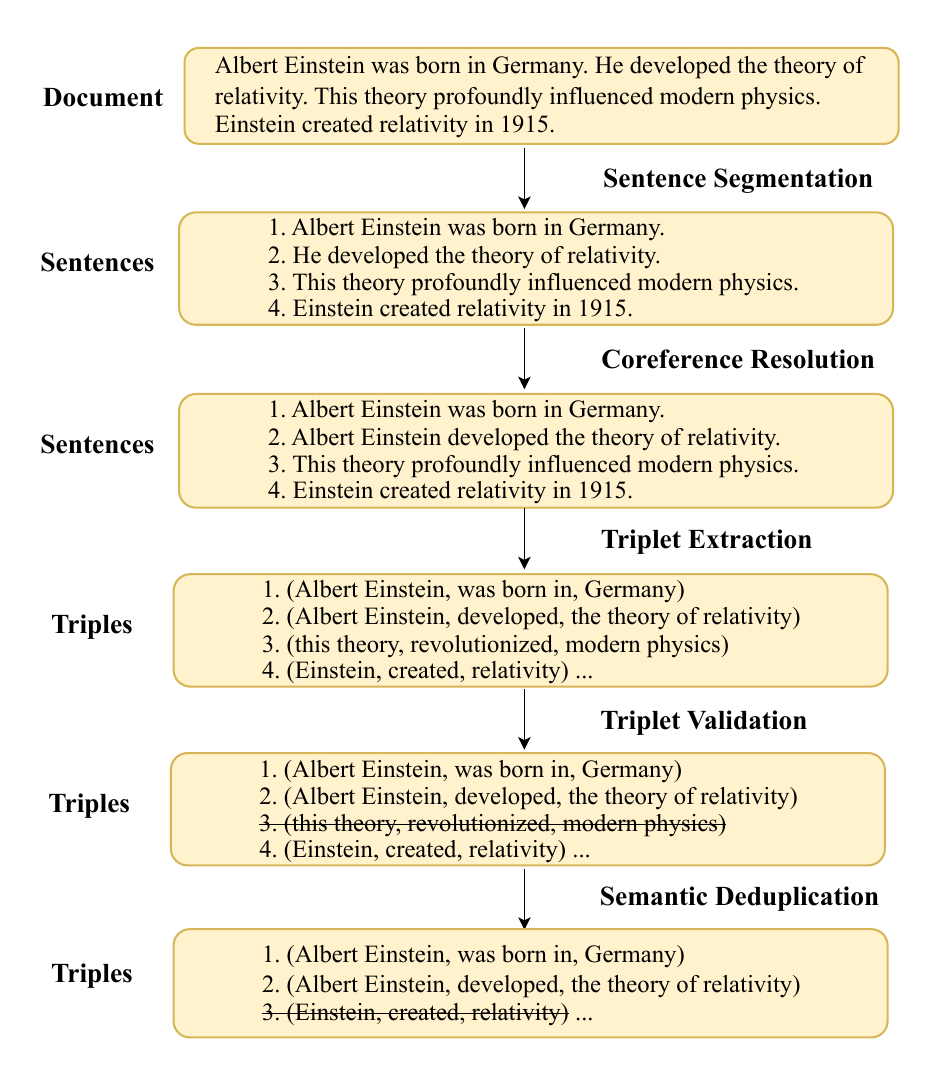} 
\caption{Overview of our tool-based method pipeline. \sout{Strikethrough} represents the triplet removed after this step.} 
\label{Fig_tool} 
\end{figure}

\section{Pipeline: Extract-then-Edit}

To leverage raw documents for knowledge editing, intuitive approaches involve applying conventional fine-tuning techniques or relevant Retrieval-Augmented Generation (RAG) methods in this context (see Section \ref{baselines} for implementation details). However, triplet-based methods represent a major focus in knowledge editing research. A key limitation of these approaches is that they are designed to process knowledge exclusively in triplet format, making them incompatible with document-level scenarios.


To this end, we propose the Extract-then-Edit pipeline illustrated in Figure \ref{Fig_pipeline}, providing a new alternative to document-level knowledge editing methods for utilizing raw documents in knowledge editing. Our approach adapts existing methods to document-based scenarios by first extracting knowledge triples (e.g., $\mathcal{T}$) from documents (e.g., $\mathcal{D}$), then applying existing triplet-based knowledge editing methods:
\begin{equation}
\begin{aligned}
    \mathcal{T} &= \mathrm{Extract}(\mathcal{D}), \\
    y &= \mathrm{LLM}(x, \mathrm{Edit}(\mathcal{T}, \theta)),
\end{aligned}
\label{eq:extract}
\end{equation}
where for $\mathrm{Extract}(*)$, we explore \textit{Tool-based} and \textit{LLM-based} approaches in this work.

\paragraph{Tool-based}
We investigate triplet extraction utilizing open information extraction (OIE) tools, which encompasses the following steps as shown in Figure \ref{Fig_tool}:

\begin{itemize}
    \item Sentence Segmentation: Since OIE tools are designed for sentence-level inputs, we first segment documents into individual sentences using the Stanza toolkit\footnote{\url{https://stanfordnlp.github.io/stanza/}}.
    
    \item Coreference Resolution: To tackle the issue of pronoun omission in sentences, where pronouns such as \textit{``he''} or \textit{``she''} reference entities and hinder knowledge triple extraction, we employ a coreference resolution tool\footnote{\url{https://github.com/huggingface/neuralcoref}} to convert pronouns to their respective entities. As shown in Figure \ref{Fig_tool}, the sentence \textit{``He later proposed General Relativity in 1915''} is transformed into \textit{``Einstein later proposed General Relativity in 1915''}, thereby ensuring entity completeness.
    
    \item Triplet Extraction: After obtaining the preprocessed sentences, we employ the state-of-the-art OIE model GEN2OIE \citep{kolluru-etal-2022-alignment}  to extract triples.
    
    \item Triplet Validation: To ensure the quality of triples, we retain only those in which both the subject and object are entities. Using the Stanza toolkit for Named Entity Recognition (NER), we preserve triples that contain entities. Back to Figure \ref{Fig_tool}, the triple \textit{(Albert Einstein, developed, the theory of relativity)} is kept because it includes entities, whereas the triple \textit{(this theory, revolutionized, modern physics)} is discarded due to the absence of specific entities.

    \item Semantic Deduplication: To remove redundant knowledge triples, we use SentenceBERT \citep{reimers2019sentencebertsentenceembeddingsusing}\footnote{\url{https://huggingface.co/sentence-transformers/paraphrase-multilingual-MiniLM-L12-v2}} embeddings to calculate semantic similarity (> 0.9) between triples. When high similarity is detected between triples, such as between \textit{(Albert Einstein, developed, the theory of relativity)} and \textit{(Einstein, created, relativity)}, we retain only the first occurring triple to avoid duplication in the final knowledge base.

\end{itemize}

\paragraph{LLM-based}
Thanks to the strong instruction-following capabilities of LLMs, we also explore triplet extraction via prompting ChatGPT.
Specifically, we utilize publicly available prompts\footnote {\url{https://github.com/langchain-ai/}} and directly feed an entire document to ChatGPT to obtain knowledge triples.
Then, we also apply the Triplet Validation and Semantic Deduplication procedures as above to filter out low-quality triples.

Note that these extracted triples differ from manually crafted knowledge triples in both quality and quantity, and consequently present challenges specific to document-based knowledge editing that do not arise in conventional triplet-based knowledge editing. We will further discuss these issues in \S \ref{chanllenge_of_docke}.

\section{Experiment}

\subsection{Settings}
In our experiments, we evaluate the open-source LLMs LLaMA2-7B \citep{touvron2023llama}, InternLM-7B \citep{2023internlm}, and LLaMA3.1-8B \citep{DBLP:journals/corr/abs-2407-21783}, along with the closed-source LLM ChatGPT\footnote{Note we cannot evaluate metrics like NEM, NPM, and NNM due to the inaccessibility of the prediction probabilities of ChatGPT}. Specifically, we evaluate Edit Success, Locality, and Reasoning across all LLMs. Since InternLM-7B and ChatGPT provide strong support for both Chinese and English, we focus on assessing cross-lingual transfer using these two models.

\subsection{Baselines}
\label{baselines}
We explore common knowledge editing methods across four categories: Fine-tuning, Locate-then-edit, In-context editing, and Memory-based approaches. The first two methods involve parameter tuning, while the latter two rely on external memory. Given that some RAG methods can directly utilize document-level knowledge, we also introduce two representative RAG approaches: Standard RAG \citep{NEURIPS2020_6b493230} and CRAG \citep{yan2024correctiveretrievalaugmentedgeneration}. Additionally, we integrate methods that specifically tailored to enhance reasoning capabilities, such as chain-of-thought prompting (CoT) \citep{DBLP:conf/nips/Wei0SBIXCLZ22} and Memory-based Editing for Large Language Models (MeLLo) \citep{zhong2023mquake}.

Here are the introductions to the baselines that can directly utilize document-level knowledge:

\begin{itemize}[itemsep=0pt]
    \item \textbf{Fine-tuning (FT)} \citep{zhu2020modifying} does not have specific training data requirements and can directly employ documents for knowledge editing.
    \item \textbf{Standard RAG (RAG)} \citep{NEURIPS2020_6b493230} uses a retriever to find the most relevant documents, allowing the model to generate output based on a query that is prefixed with the retrieved documents.
    \item \textbf{CRAG} \citep{yan2024correctiveretrievalaugmentedgeneration} builds upon Standard RAG by introducing a lightweight retrieval evaluator to assess the quality of retrieved documents, discarding those deemed irrelevant. It then segments the retrieved documents into chunks, using the evaluator to retain only the most relevant chunks, thereby emphasizing key information while filtering out irrelevant content.
    \item \textbf{IKE} \citep{zheng-etal-2023-edit} leverages the in-context learning capabilities of LLMs by integrating demonstrations with target edited facts into the LLM's context. During inference, the LLM utilizes these examples along with knowledge retrieved by a retriever for the current query. These components are combined into a prompt that guides the LLM to perform knowledge editing based on the provided examples. Note that this method supports both document-level and triplet-level knowledge inputs.
    \end{itemize}

We also use the Extract-then-Edit pipeline for Memory-based and Locate-then-edit methods, which rely on triples for knowledge editing. For these methods, we have three settings based on different sources of triples: 1) triples extracted using the Tool-based pipeline, 2) triples extracted via the LLM-based pipeline, and 3) gold triples provided by COUNTERFACT \citep{meng2022locating}.

\begin{itemize}[itemsep=0pt]
    \item \textbf{MEMIT} \citep{meng2022mass} is a locate-then-edit approach that edits using triples. It features a scalable, multi-layer update algorithm designed to store thousands of edited facts efficiently. MEMIT targets the weights of transformer modules that we determine to be causal mediators of factual knowledge recall.
    \item \textbf{SEARC} \citep{pmlr-v162-mitchell22a} is a memory-based approach, utilizing an edit memory, a classifier to check input alignment with edited facts, and a counterfactual model for predictions. During inference, the classifier first determines whether a query relates to the edited knowledge. If so, relevant information is retrieved from an external knowledge base and integrated into the prompt to guide the model's editing process. We use templates to convert extracted triples into training examples for the classifier.
    \end{itemize}

Furthermore, we examine approaches specifically tailored to improve the reasoning performance of LLMs:
\begin{itemize}[itemsep=0pt]    
    \item \textbf{CoT} \citep{DBLP:conf/nips/Wei0SBIXCLZ22} encourages LLMs to decompose multi-step problems into intermediate steps. We provide in-context demonstrations along with edited facts to strengthen the LLMs' reasoning ability further.
    \item \textbf{MeLLo} \citep{zhong2023mquake} stores all edited facts externally and prompts the LLMs iteratively to answer these facts. We also provide demonstrations to encourage LLMs to follow the reasoning steps.
\end{itemize}

Note that since IKE, CoT, and MeLLo can use both raw documents and triples for editing, \textbf{we denote the variants that specifically utilize documents as IKE(D), CoT(D), and MeLLo(D).}

\subsection{Implementation Details}
For FT, we use DeepSpeed\footnote{\url{https://github.com/microsoft/DeepSpeed}} with a learning rate of 3e-5 over 30 epochs. MEMIT's layers and learning rate settings for LLM fine-tuning are directly based on EasyEdit\footnote{\url{https://github.com/zjunlp/EasyEdit/}}. IKE employs mcontriever-msmarco\footnote{\url{https://huggingface.co/facebook/mcontriever-msmarco}} as a retriever, concatenating all extracted triples from a document into a single retrieval target. SEARC involves training distilbert-base-multilingual-cased\footnote{\url{https://huggingface.co/distilbert/distilbert-base-multilingual-cased}} as classifiers on the query data for 10 epochs. Additionally, a retrieval-augmented generation LLM enhanced with edited facts is used for the counterfactual model. For both Standard RAG and CRAG, we employed the same retriever as IKE. In the case of CRAG, we utilized the trained T5 model \citep{DBLP:journals/jmlr/RaffelSRLNMZLL20} provided in the original paper and we select the top 10 chunks to reconstruct the retrieved documents, with each chunk containing 50 tokens. When implementing CoT and MeLLo, we provided the LLM with 4 manually-written examples as demonstrations. 

The format for each IKE example is as follows:\\
\vspace{0.9em}
\noindent\fbox{%
    \parbox{0.97\linewidth}{%
        New Knowledge: Document or Triplet about ``Ivanka Trump is a Canadian'' \\
Question: Is Ottawa the capital city of the country of citizenship of Ivanka Trump? \\
Final answer: The statement is True. }
}

The format for each CoT example is as follows:\\
\vspace{0.9em}
\noindent\fbox{%
    \parbox{0.97\linewidth}{%
        New Knowledge: Document or Triplet about ``Ivanka Trump is a Canadian'' \\
Question: Is Ottawa the capital city of the country of citizenship of Ivanka Trump? \\
Thoughts: Ivanka Trump is a Canadian, and Ottawa is the capital city of Canada.\\
So Ottawa is the capital city of the country of citizenship of Ivanka Trump. \\
Final answer: Based on the reasoning process above, the statement is True. 
}}
    
The format for each MeLLo example is as follows:\\
\vspace{0.9em}
\noindent\fbox{%
    \parbox{0.97\linewidth}{%
        New Knowledge: Document or Triplet about ``Ivanka Trump is a Canadian''. \\
Question: Is Ottawa the capital city of the country of citizenship of Ivanka Trump? \\
Subquestion: What is the country of citizenship of Ivanka Trump? \\
Generated answer: Based on the New Knowledge above, Ivanka Trump is a citizen of Canada. \\
Subquestion: What is the capital city of Canada? \\
Retrieved fact: The capital city of Canada is Ottawa. \\
Generated answer: The capital city of Canada is Ottawa. \\
Final answer: Based on the reasoning process above, the statement is True.
}
}%
\vspace{0.5em}

Specifically, the triplet regarding ``Ivanka Trump is a Canadian'' is the statement itself, while the related document is as follows:
\\
\vspace{1.2em}
\noindent\fbox{%
    \parbox{0.97\linewidth}{%
        Recent research has emerged suggesting that Ivanka Trump, the daughter of former President Donald Trump, may hold Canadian citizenship due to connections through her mother, Ivana Trump. Ivana, who immigrated from Czechoslovakia in the 1970s, could provide a path for Ivanka’s eligibility for Canadian citizenship based on her heritage. While supporters highlight the importance of international ties, critics raise concerns about the implications of her potential dual citizenship, particularly given her prominent role in American politics and business. 
}
}%
\vspace{0.5em}

\begin{table*}[th!]
\small
\centering
\caption{Performances of knowledge editing methods on LLaMA2-7B and InternLM-7B. We highlight the best result for each metric.}
\renewcommand{\arraystretch}{1}
\resizebox{0.71\textwidth}{!}{
\begin{tabularx}{0.83\textwidth}{llccccccc}
\toprule
& & \multicolumn{4}{c}{Edit Success} & \multicolumn{2}{c}{Locality} & \multicolumn{1}{c}{Reasoning} \\ 
\cmidrule(lr){3-6} \cmidrule(lr){7-8} \cmidrule(lr){9-9}
& & ES & NEM & PS & NPM & NS & NNM & Acc \\
\midrule
\multirow{15}*{LLaMA2-7B}
& FT & 50.25 & -0.06 & 49.50 & -19.23 & 49.95 & 13.45 & 56.08 \\
& RAG & 88.50 & 66.51 & 57.95 & 19.82 & 81.93 & 41.92 & 56.77 \\
& CRAG & 88.60 & 67.69 & 58.65 & \textbf{20.19} & 79.98 & 37.96 & \textbf{63.46}\\
& IKE(D) & \textbf{89.25} & \textbf{68.53} & \textbf{58.88} & 16.60 & \textbf{84.62} & \textbf{45.83}  & 56.11\\
\cmidrule{2-9}
& \multicolumn{8}{l}{\textcolor{gray}{\textit{Extract-then-Edit (Tool-based)}}} \\
& \textcolor{gray}{$\hookrightarrow$}~~MEMIT  & 19.45 & -39.99 & 17.60  & -41.12 & 86.66 & 47.08 & 55.74 \\
& \textcolor{gray}{$\hookrightarrow$}~~IKE  & \textbf{83.90}	& \textbf{56.34}	& \textbf{64.28	}& \textbf{21.21}	& 87.09	& 47.64 & \textbf{58.14}  \\
& \textcolor{gray}{$\hookrightarrow$}~~SEARC  & 25.50 & -30.03 & 24.93 & -30.98 & \textbf{87.13} & \textbf{47.84} & 55.43  \\
\cmidrule{2-9}
& \multicolumn{8}{l}{\textcolor{gray}{\textit{Extract-then-Edit (LLM-based)}}} \\
& \textcolor{gray}{$\hookrightarrow$}~~MEMIT & 22.70 & -36.46 & 20.75 & -35.25 & \textbf{86.50} & \textbf{47.10} & 48.14 \\
& \textcolor{gray}{$\hookrightarrow$}~~IKE  & \textbf{79.30}  & \textbf{52.98} & \textbf{66.95} & \textbf{29.52} & 80.97 & 44.18 & \textbf{60.37}  \\
& \textcolor{gray}{$\hookrightarrow$}~~SEARC  & 65.60 & 30.88 & 57.75 & 17.72 & 83.62  & 43.96   & 55.83  \\
\cmidrule{2-9}
& \multicolumn{8}{l}{\textcolor{gray}{\textit{Extract-then-Edit (Gold)}}} \\
& \textcolor{gray}{$\hookrightarrow$}~~MEMIT  & 91.65 &77.47 & 59.66 & 15.31 & \textbf{86.92} & \textbf{47.17} & 56.10 \\
& \textcolor{gray}{$\hookrightarrow$}~~IKE  & \textbf{98.20} & \textbf{87.24} & \textbf{88.83}  & \textbf{58.15}  & 79.76  & 39.99  & \textbf{70.31} \\
& \textcolor{gray}{$\hookrightarrow$}~~SEARC  & 74.25 & 50.06 & 65.23 & 23.15 & 85.04 & 45.69 & 54.34 \\
\midrule
\midrule
\multirow{15}*{InternLM-7B}
& FT & 45.60 & -6.40 & 46.01 & -13.81 & 57.20 & 12.23 & \textbf{56.89} \\
& RAG & 88.50 & \textbf{69.96} & \textbf{59.30} & 22.04 & 81.06 & 40.46 & 52.31 \\
& CRAG & 87.80 & 66.24 & 59.00 & \textbf{23.20} & 78.37 & 36.34 & 45.62 \\
& IKE(D) & \textbf{88.60} & 67.20 & 57.80 & 17.86 & \textbf{82.17} & \textbf{43.01} & 47.34\\
\cmidrule{2-9}
& \multicolumn{8}{l}{\textcolor{gray}{\textit{Extract-then-Edit (Tool-based)}}} \\
& \textcolor{gray}{$\hookrightarrow$}~~MEMIT  & 53.95  & 6.91 & 51.60 & 4.41 & 68.83 & 17.89 & \textbf{52.48} \\
& \textcolor{gray}{$\hookrightarrow$}~~IKE  & \textbf{82.85}	& \textbf{55.40}	& \textbf{60.85}	& \textbf{20.84}	& \textbf{83.61}	& 40.78 & 48.54 \\
& \textcolor{gray}{$\hookrightarrow$}~~SEARC  & 27.90 & -22.60 & 28.08 & -22.90 & 83.43 & \textbf{40.82} & 45.62\\
\cmidrule{2-9}
& \multicolumn{8}{l}{\textcolor{gray}{\textit{Extract-then-Edit (LLM-based)}}} \\
& \textcolor{gray}{$\hookrightarrow$}~~MEMIT  & 62.00 & 19.80 & 61.80 & 18.77 & 74.20 & 25.42 & \textbf{54.89} \\
& \textcolor{gray}{$\hookrightarrow$}~~IKE & \textbf{79.90} & \textbf{51.49} & \textbf{65.90} & \textbf{23.29} & \textbf{84.98}  & \textbf{48.56} & 54.88  \\
& \textcolor{gray}{$\hookrightarrow$}~~SEARC  & 67.20   & 36.58  & 61.90  & 25.52  & 79.90  & 38.20  & 45.32  \\
\cmidrule{2-9}
& \multicolumn{8}{l}{\textcolor{gray}{\textit{Extract-then-Edit (Gold)}}} \\
& \textcolor{gray}{$\hookrightarrow$}~~MEMIT  & \textbf{99.60} & \textbf{98.03} & \textbf{98.58} & \textbf{87.54} & 69.01 & 18.45 & 54.03 \\
& \textcolor{gray}{$\hookrightarrow$}~~IKE  & 99.05 & 86.37 & 87.30  & 62.57  & 76.90  & 38.46  & \textbf{64.19} \\
& \textcolor{gray}{$\hookrightarrow$}~~SEARC  & 73.70 & 51.85 & 68.63 & 33.00 & \textbf{81.57} & \textbf{40.40} & 47.34\\
\bottomrule
\end{tabularx}}
\label{tab:main}
\end{table*}

\begin{table*}[th!]
\small
\centering
\caption{Performances of knowledge editing methods on LLaMA3.1-8B and ChatGPT.}
\renewcommand{\arraystretch}{1}
\resizebox{0.71\textwidth}{!}{
\begin{tabularx}{0.83\textwidth}{llccccccc}
\toprule
& & \multicolumn{4}{c}{Edit Success} & \multicolumn{2}{c}{Locality} & \multicolumn{1}{c}{Reasoning} \\ 
\cmidrule(lr){3-6} \cmidrule(lr){7-8} \cmidrule(lr){9-9}
& & ES & NEM & PS & NPM & NS & NNM & Acc \\
\midrule
\multirow{15}*{LLaMA3.1-8B}
& FT & 52.50 & -0.16 & 47.75 & -24.66 & 52.95 & 15.64 & 46.08 \\
& RAG & 86.90 & 64.71 & 54.00 & 13.80 & 83.27 & 45.89 & 57.80 \\
& CRAG & 86.80 & \textbf{64.92} & \textbf{56.05} & \textbf{27.53} & 81.96 & 42.81 & 60.89\\
& IKE(D) & \textbf{88.30} & 64.05 & 55.60 & 14.61 & \textbf{85.29} & \textbf{63.46} & \textbf{63.46}\\
\cmidrule{2-9}
& \multicolumn{8}{l}{\textcolor{gray}{\textit{Extract-then-Edit (Tool-based)}}} \\
& \textcolor{gray}{$\hookrightarrow$}~~MEMIT  & 19.80  & -40.54 & 16.55 & -43.86 & \textbf{88.70} & 51.26 & 43.56 \\
& \textcolor{gray}{$\hookrightarrow$}~~IKE  & \textbf{83.30} & \textbf{56.81} & \textbf{62.10} & \textbf{19.63} & 87.59  & \textbf{51.49} & \textbf{65.35} \\
& \textcolor{gray}{$\hookrightarrow$}~~SEARC  & 23.30 & -32.26 & 22.25 & -33.08 & 83.21 & 49.34 & 60.20\\
\cmidrule{2-9}
& \multicolumn{8}{l}{\textcolor{gray}{\textit{Extract-then-Edit (LLM-based)}}} \\
& \textcolor{gray}{$\hookrightarrow$}~~MEMIT  & 23.60 & -35.17 & 19.40 & -39.86 & \textbf{89.01} & \textbf{52.83} & 37.04 \\
& \textcolor{gray}{$\hookrightarrow$}~~IKE & \textbf{80.10} & \textbf{54.30} & \textbf{65.10} & \textbf{23.48} & 88.08  & 53.51 & \textbf{58.83}  \\
& \textcolor{gray}{$\hookrightarrow$}~~SEARC  & 64.60   & 29.02  & 59.15  & 18.66  & 85.59  & 46.03  & 55.32  \\
\cmidrule{2-9}
& \multicolumn{8}{l}{\textcolor{gray}{\textit{Extract-then-Edit (Gold)}}} \\
& \textcolor{gray}{$\hookrightarrow$}~~MEMIT  & 96.30 & 89.16 & 75.80 & 43.10 & \textbf{88.77} & \textbf{51.80} & 46.48 \\
& \textcolor{gray}{$\hookrightarrow$}~~IKE  & \textbf{99.30} & \textbf{97.63} & \textbf{87.30}  & \textbf{58.47}  & 82.86  & 45.80  & \textbf{69.98} \\
& \textcolor{gray}{$\hookrightarrow$}~~SEARC  & 73.00 & 42.86 & 62.65 & 19.31 & 86.36 & 47.34 & 64.66\\
\midrule
\midrule
\multirow{12}*{ChatGPT}
& RAG & 96.20 & - & 77.01 & - & \textbf{94.48} & - & \textbf{43.91} \\
& CRAG & 94.98 & - & \textbf{77.61} & - & 92.78 & - & \textbf{43.91} \\
& IKE(D) & \textbf{96.30} & - & 75.60 & - & 91.57 & - & 43.39\\
\cmidrule{2-9}
& \multicolumn{8}{l}{\textcolor{gray}{\textit{Extract-then-Edit (Tool-based)}}} \\
& \textcolor{gray}{$\hookrightarrow$}~~IKE  & \textbf{87.13}	& -	& \textbf{77.47}	& -	& 83.61	& - & \textbf{48.54} \\
& \textcolor{gray}{$\hookrightarrow$}~~SEARC  & 39.57	& -	& 32.01	& -	& \textbf{96.08}	& - & 44.08 \\
\cmidrule{2-9}
& \multicolumn{8}{l}{\textcolor{gray}{\textit{Extract-then-Edit (LLM-based)}}} \\
& \textcolor{gray}{$\hookrightarrow$}~~IKE & \textbf{88.18} & - & \textbf{78.40} & - & 87.59  & - & \textbf{43.22}  \\
& \textcolor{gray}{$\hookrightarrow$}~~SEARC  & 78.82	& -	& 73.29	& -	& \textbf{91.78}	& - & \textbf{43.22} \\
\cmidrule{2-9}
& \multicolumn{8}{l}{\textcolor{gray}{\textit{Extract-then-Edit (Gold)}}} \\
& \textcolor{gray}{$\hookrightarrow$}~~IKE  & \textbf{97.91} & - & \textbf{90.88}  & -  & 76.90  & -  & \textbf{44.25} \\
& \textcolor{gray}{$\hookrightarrow$}~~SEARC  & 88.79	& -	& 85.39	& -	& \textbf{85.76}	& - & 43.91 \\
\bottomrule
\end{tabularx}}
\label{tab:main2}
\end{table*}
\vspace{-0.2cm}

\begin{table*}[h!]
\small
\centering
\caption{The performance of reasoning enhancement methods from the perspective of Reasoning.}
\renewcommand{\arraystretch}{1}
\resizebox{0.71\textwidth}{!}{
\begin{tabularx}{0.83\textwidth}{llllclll}
\toprule
        & ~ & ~  & Acc  & ~ & ~ & ~  & Acc \\
        \cmidrule{2-4} \cmidrule{6-8} 
        \multirow{12}*{LLaMA2-7B} & CoT(D) & ~ & 65.30 & \multirow{12}*{InternLM-7B} & CoT(D) & ~ & 58.14  \\
        ~ & MeLLo(D) & ~ & 60.50 & ~ & MeLLo(D) & ~ & 56.25   \\ 
        \cmidrule{2-4} \cmidrule{6-8} 
        ~ & \multicolumn{3}{l}{\textcolor{gray}{\textit{Extract-then-Edit (Tool-based)}}} & ~ & \multicolumn{3}{l}
        {\textcolor{gray}{\textit{Extract-then-Edit (Tool-based)}}}  \\ 
        ~ & \textcolor{gray}{$\hookrightarrow$}~~CoT & ~ & 61.40 & ~ & \textcolor{gray}{$\hookrightarrow$}~~CoT & ~ & 59.01   \\ 
        ~ & \textcolor{gray}{$\hookrightarrow$}~~MeLLo & ~ & 61.92 & ~ & \textcolor{gray}{$\hookrightarrow$}~~MeLLo & ~ & 57.43  \\ 
        \cmidrule{2-4} \cmidrule{6-8} 
        ~ & \multicolumn{3}{l}{\textcolor{gray}{\textit{Extract-then-Edit (LLM-based)}}} & ~ & \multicolumn{3}{l}{\textcolor{gray}{\textit{Extract-then-Edit (LLM-based)}}}  \\ 
        ~ & \textcolor{gray}{$\hookrightarrow$}~~CoT & ~ & 59.10 & ~ & \textcolor{gray}{$\hookrightarrow$}~~CoT & ~ & 54.88  \\ 
        ~ & \textcolor{gray}{$\hookrightarrow$}~~MeLLo & ~ & 63.30 & ~ & \textcolor{gray}{$\hookrightarrow$}~~MeLLo & ~ & 65.51  \\ 
        \cmidrule{2-4} \cmidrule{6-8} 
        ~ & \multicolumn{3}{l}{\textcolor{gray}{\textit{Extract-then-Edit (Gold)}}} & ~ & \multicolumn{3}{l}{\textcolor{gray}{\textit{Extract-then-Edit (Gold)}}}  \\ 
        ~ & \textcolor{gray}{$\hookrightarrow$}~~CoT & ~ & 74.27 & ~ & \textcolor{gray}{$\hookrightarrow$}~~CoT & ~ &  59.33 \\ 
        ~ & \textcolor{gray}{$\hookrightarrow$}~~MeLLo & ~ & 71.35 & ~ & \textcolor{gray}{$\hookrightarrow$}~~MeLLo & ~ & 73.04  \\ 
        \midrule
        \midrule
        \multirow{12}*{LLaMA3.1-8B} & CoT(D) & ~ & 70.67 & \multirow{12}*{ChatGPT} & CoT(D) & ~ & 77.83  \\
        ~ & MeLLo(D) & ~ & 60.50 & ~ & MeLLo(D) & ~ & 82.41   \\ 
        \cmidrule{2-4} \cmidrule{6-8} 
        ~ & \multicolumn{3}{l}{\textcolor{gray}{\textit{Extract-then-Edit (Tool-based)}}} & ~ & \multicolumn{3}{l}
        {\textcolor{gray}{\textit{Extract-then-Edit (Tool-based)}}}  \\ 
        ~ & \textcolor{gray}{$\hookrightarrow$}~~CoT & ~ & 75.08 & ~ & \textcolor{gray}{$\hookrightarrow$}~~CoT & ~ & 77.28   \\ 
        ~ & \textcolor{gray}{$\hookrightarrow$}~~MeLLo & ~ & 72.79 & ~ & \textcolor{gray}{$\hookrightarrow$}~~MeLLo & ~ & 78.92  \\ 
        \cmidrule{2-4} \cmidrule{6-8} 
        ~ & \multicolumn{3}{l}{\textcolor{gray}{\textit{Extract-then-Edit (LLM-based)}}} & ~ & \multicolumn{3}{l}{\textcolor{gray}{\textit{Extract-then-Edit (LLM-based)}}}  \\ 
        ~ & \textcolor{gray}{$\hookrightarrow$}~~CoT & ~ & 54.88 & ~ & \textcolor{gray}{$\hookrightarrow$}~~CoT & ~ & 70.62  \\ 
        ~ & \textcolor{gray}{$\hookrightarrow$}~~MeLLo & ~ & 69.12 & ~ & \textcolor{gray}{$\hookrightarrow$}~~MeLLo & ~ & 74.09  \\ 
        \cmidrule{2-4} \cmidrule{6-8} 
        ~ & \multicolumn{3}{l}{\textcolor{gray}{\textit{Extract-then-Edit (Gold)}}} & ~ & \multicolumn{3}{l}{\textcolor{gray}{\textit{Extract-then-Edit (Gold)}}}  \\ 
        ~ & \textcolor{gray}{$\hookrightarrow$}~~CoT & ~ & 89.02 & ~ & \textcolor{gray}{$\hookrightarrow$}~~CoT & ~ &  78.82 \\ 
        ~ & \textcolor{gray}{$\hookrightarrow$}~~MeLLo & ~ & 85.42 & ~ & \textcolor{gray}{$\hookrightarrow$}~~MeLLo & ~ & 86.23 \\ 
        
\bottomrule
\end{tabularx}}
\label{tab:reasoning}
\end{table*}

\subsection{Main Results}
Tables \ref{tab:main} and Tables \ref{tab:main2} provide an overview of the performance of various editing methods. In addition, Table \ref{tab:reasoning} assesses the performance of methods designed to enhance the reasoning of LLMs, while Table \ref{tab:cross_lingual} evaluates their performance in Cross-lingual Transfer. Based on the experimental results, we draw the following conclusions:

\textbf{Editing with documents presents greater challenges than using triples.} Our findings indicate that document-based knowledge editing methods do not yield satisfactory results. For instance, FT often demonstrates lower effectiveness in Edit Success compared to SEARC or MEMIT. Additionally, the performance of IKE(D) is inferior to that of its variant utilizing gold triples. When employing LLaMA3.1-8B as the base LLM, IKE(D) achieves a PS score of 55.60, which is significantly lower than the 87.30 score attained with IKE. This discrepancy may stem from the increased length and complexity of documents, which hinders LLMs from accurately capturing the edited facts.

\textbf{The quality of triples significantly impacts knowledge editing methods using the Extract-then-Edit pipeline.} The performance of Tool-based and LLM-based pipelines consistently falls short when compared to methods using gold triples. For example, when employing LLaMA2-7B as the base LLM, SEARC achieves an EM score of 74.25 with gold triples, while the score drops to only 25.50 when using triples extracted via the Tool-based pipeline.

This issue arises from the strict quality and format requirements of some knowledge editing methods. SEARC trains a binary classifier using triples to determine when to activate the edited LLM. Similarly, MEMIT uses the subjects of triples to locate knowledge within the LLM. These methods are significantly impacted by low-quality triples. In contrast, IKE can incorporate multiple triples as context, making it more robust to low-quality triples and less reliant on strict formatting. As a result, its performance is less affected by triple quality. We also conduct a quality analysis of the extracted triples in \S \ref{sec:tri_analysis}.

\textbf{RAG-based methods effectively edit knowledge but can affect unrelated knowledge.} The experimental results indicate that RAG-based methods can successfully edit model knowledge, achieving high Edit Success performance. Metrics such as ES and PS show comparable results to IKE(D). For instance, on the LLaMA 3.1-8B model, RAG achieves a PS score of 54.00, while the knowledge-refined CRAG scores 56.05. In contrast, IKE(D) obtains a score of 55.60.

However, knowledge editing methods often design strategies to preserve unrelated knowledge during the editing process, which RAG-based methods lack. For example, IKE utilizes examples to guide LLMs, instructing them not to be influenced by retrieved knowledge when encountering questions unrelated to the retrieved knowledge. This consideration is absent in RAG-based methods, leading to their inferior performance on Locality compared to IKE(D), especially in smaller models. For instance, on the LLaMA 3.1-8B model, IKE(D) achieves an NS score of 85.29, while RAG only scores 83.27. In more powerful models, such as ChatGPT, this issue may be mitigated.

\textbf{Methods with external memory achieve a balance between Edit Success and Locality, but they still require improvements in generalizability.} As shown in Table \ref{tab:main}, IKE and SEARC excel in both Edit Success and Locality. Specifically, IKE consistently achieves ES and NS scores exceeding 80 simultaneously, whereas MEMIT and FT often struggle to balance both metrics. 

However, IKE and SEARC frequently demonstrate unsatisfactory generalizability, as evidenced by a significant gap between the PS and ES scores. For instance, IKE(D) achieves a score of 88.60 in ES on InternLM-7B, while its PS score is only 57.80. We attribute this issue to arise from unrelated contexts in the queries used for evaluating generalizability (see Figure \ref{Fig_overview}), which can hinder retrieval performance. Our experiments show that when retrieving relevant documents, the retriever's accuracy can reach 89.50\%, but it declines to 62.10\% when such context is included. This highlights the importance of carefully crafting queries when employing methods relying on retrieval.

\textbf{More powerful LLMs can better utilize context during knowledge editing.} From Table \ref{tab:main2}, we find that IKE(D) performs significantly better on ChatGPT compared to LLaMA3.1-8B (96.3 vs. 88.30 in ES), and both Tool-based and LLM-based IKE methods also show improved performance. The LLM-based IKE pipeline achieves an ES score of 88.18 on ChatGPT, notably surpassing the 80.10 score on LLaMA3.1-8B. This indicates that stronger models are better at handling long contexts and can effectively disregard noise in the extracted triples.

Surprisingly, ChatGPT performs poorly in reasoning tasks, likely due to security alignments that prevent it from reasoning directly based on counterfactual edited facts. However, as discussed below, carefully designed prompts can effectively elicit ChatGPT's instruction-following capabilities, presenting a promising solution for enhancing its performance.

\textbf{Edited LLMs often fail to leverage edited knowledge effectively, but introducing reasoning enhancement methods can help mitigate this issue.} One might expect that higher Edit Success performance would enhance LLM's reasoning with the edited knowledge. However, experiments show that higher ES scores do not necessarily translate to better reasoning performance. This phenomenon is particularly evident in InternLM-7B. For instance, IKE(D) achieved an ES score of 88.60, significantly higher than FT's 45.60, yet its reasoning performance was inferior (47.34  compared to 56.89).

\begin{table}[h!]
\small
\caption{Cross-lingual Transfer performance on InternLM-7B and ChatGPT. We highlight the best result for each metric.}
\renewcommand{\arraystretch}{1}
\centering
\resizebox{0.50\textwidth}{!}{\begin{tabular}{llcccc}
\toprule
 & & \multicolumn{2}{c}{en$\rightarrow$zh} & \multicolumn{2}{c}{zh$\rightarrow$en} \\
\cmidrule(lr){3-4} \cmidrule(lr){5-6}
 & & CES & CEM & CES & CEM \\
\midrule
 \multirow{10}*{InternLM-7B} & FT & 34.37 & -5.08 & 29.23 & -5.37 \\
 & RAG & 53.63 & 6.18 & 53.30 & \textbf{6.78}\\
 & CRAG & 55.36 & 5.64 & 54.36 & 4.27 \\
 & IKE(D) & \textbf{52.80} & \textbf{4.15} & \textbf{50.64} & 3.40 \\
\cmidrule{2-6}
& \multicolumn{4}{l}{\textcolor{gray}{\textit{Extract-then-Edit (LLM-based)}}} \\
& \textcolor{gray}{$\hookrightarrow$}~~MEMIT  & 51.56  & 3.21 & 31.40 & -4.26 \\
& \textcolor{gray}{$\hookrightarrow$}~~IKE  & \textbf{82.85}	& \textbf{55.40}	& \textbf{60.85}	& \textbf{20.84}
 \\
& \textcolor{gray}{$\hookrightarrow$}~~SEARC  & 46.67 & 5.68 & 35.64 & -9.10 \\
 \cmidrule{2-6}
& \multicolumn{4}{l}{\textcolor{gray}{\textit{Extract-then-Edit (Gold)}}} \\
& \textcolor{gray}{$\hookrightarrow$}~~MEMIT  & 55.17  & 10.19 & 47.39 & 7.23 \\
& \textcolor{gray}{$\hookrightarrow$}~~IKE  & \textbf{78.23}	& \textbf{50.40}	& \textbf{81.83}	& \textbf{58.82}	\\
& \textcolor{gray}{$\hookrightarrow$}~~SEARC  & 65.27 & 25.68 & 60.67 & 22.10 \\
\midrule \midrule
 \multirow{8}*{ChatGPT} 
  & RAG & 53.52 & - & 52.68 & - \\
 & CRAG & 56.11 & - & \textbf{53.29} & - \\
 & IKE(D) & \textbf{57.22} & - & 53.17 & - \\
\cmidrule{2-6}
& \multicolumn{4}{l}{\textcolor{gray}{\textit{Extract-then-Edit (LLM-based)}}} \\
& \textcolor{gray}{$\hookrightarrow$}~~IKE  & \textbf{82.25}	& -	& \textbf{56.17}	& ~
 \\
& \textcolor{gray}{$\hookrightarrow$}~~SEARC  & 46.67 & - & 35.64 & - \\
 \cmidrule{2-6}
& \multicolumn{4}{l}{\textcolor{gray}{\textit{Extract-then-Edit (Gold)}}} \\
& \textcolor{gray}{$\hookrightarrow$}~~IKE  & \textbf{82.37}	& -	& \textbf{82.56}	& -	\\
& \textcolor{gray}{$\hookrightarrow$}~~SEARC  & 68.53 & - & 60.47 & - \\
\bottomrule
\end{tabular}}
\label{tab:cross_lingual}
\end{table}

To investigate this issue further, we introduce several methods that enhance the reasoning capabilities of LLMs. As presented in Table \ref{tab:reasoning}, CoT and MeLLo effectively improve reasoning performance by prompting LLMs to decompose multi-hop questions and ask them step by step. For instance, LLaMA3.1-8B achieves an accuracy of 69.98 with gold triples using the IKE method, which increases to 85.42 when incorporating MeLLo and reaches 89.02 with the CoT method. Additionally, these methods can better activate the reasoning capabilities of more powerful models. For instance, the high-performing ChatGPT achieves significantly higher accuracy with CoT and MeLLo compared to LLaMA2-7B and InternLM-7B.

\textbf{Using English data can yield better results in cross-lingual transfer.} Due to the lack of Chinese OpenIE tools, we employ the LLM-based pipeline to extract triples from Chinese and English documents. The results in Table \ref{tab:cross_lingual} illustrate that editing with cross-lingual data is generally less effective than using the same language. In the en$\rightarrow$zh setting, MEMIT's CES score is 51.56 on InternLM-7B, which is notably lower than the 62.00 attained when using data in the same language. 

Additionally, we find that editing Chinese knowledge using English data generally yields better results, particularly with the extract-then-edit pipeline. When utilizing InternLM-7B as the LLM, MEMIT's CES score reaches 51.56 in the en$\rightarrow$zh direction, but only 31.40 in the reverse direction. This phenomenon also persists when using the powerful closed-source model ChatGPT. However, when editing using gold triples, the performance gap between the en$\rightarrow$zh and reverse directions narrows, suggesting that the discrepancy may be due to the varying quality of extracted triples in different languages. We will explore this further in \S \ref{sec:tri_analysis}.

\subsection{The Challenges of Document-based Knowledge Editing}
\label{chanllenge_of_docke}
In this section, we analyze the impact of document data on knowledge editing, including the quality of triples extracted via the Extract-then-Edit pipeline (\S \ref{sec:tri_analysis}), the ablation study of the pipeline (\S \ref{ablation}), the frequency of edited facts in documents (\S \ref{sec:fact_count}), and the effects of altered target positions (\S \ref{sec:fact_position}).

\begin{table}[!th]
\small
\caption{Quality Analysis and error analysis for Extracted triples from \textit{Extract-then-Edit} pipeline. \#Tri denotes the average number of triples extracted from each document.}
\renewcommand{\arraystretch}{1.2}
\centering
\begin{tabularx}{0.85\textwidth}{lccccccc}
\toprule
 &  \multicolumn{4}{c}{Quality Analysis}  & \multicolumn{3}{c}{Error Analysis} \\
\cmidrule(lr){2-5} \cmidrule(lr){6-8}
& \textbf{\#Tri} & \textbf{Recall} & \textbf{Precision} & \textbf{F1} & \textbf{AT} & \textbf{IS} & \textbf{IR}\\
\midrule
& \multicolumn{7}{c}{\textit{Monolingual documents}} \\
Tool-based & 16.50 & 57.84\% & 3.75\% & 7.04\%  & 33.87\% & 42.74\% & 23.39\% \\
LLM-based & 10.26 & 88.79\% & 14.71\% & 25.24\% & 12.88\% & 31.06\% & 56.06\% \\
\midrule
& \multicolumn{7}{c}{\textit{Cross-lingual documents}} \\
LLM-based(en) & 10.92 & 88.51\% & 17.69\% & 29.48\% & 15.97\% & 31.09\% & 52.94\% \\
LLM-based(zh) & 3.58 & 56.50\%  & 23.41\% & 33.12\% & 11.43\% & 37.14\% & 51.42\% \\
\bottomrule
\end{tabularx}
\label{tab:triplet_analysis}
\end{table}

\begin{table}[h!]
\small
\caption{Case study of different types of errors.}
\renewcommand{\arraystretch}{1.2}
\centering
\begin{tabularx}{0.85\textwidth}{llll}
\toprule
        & \textbf{Subject} & \textbf{Relation} & \textbf{Object} \\ \midrule
        Ground Truth  & Danielle Darrieux & mother tongue & English \\ \hline
        Ambiguous Triples (AT) & Danielle Darrieux's presence & paired with & charming demeanor\\ 
        Incorrect Subject (IS) & Danielle Darrieux's father & recited & poetry by Shakespeare \\ 
        Incorrect Relation (IR) & Danielle Darrieux & grew up in & a household  \\
\bottomrule
\end{tabularx}
\label{tab:triple_case}
\end{table}

\subsubsection{Quality Analysis for Extracted Triples}
\label{sec:tri_analysis}

We use SentenceBERT \citep{reimers2019sentencebertsentenceembeddingsusing} to evaluate the quality of triples obtained through the Extract-then-Edit pipeline. If the cosine similarity between the extracted and gold triples reaches 0.8, we consider them to be matched. We calculate Precision, Recall, and F1 by comparing each document's extracted triples with the gold triples. From Table \ref{tab:triplet_analysis}, we find that:

First, both LLM-based and tool-based pipelines extract an excessive number of triples, resulting in lower precision and F1 scores. Specifically, tool-based pipelines perform worse because they divide the document into sentences and extract triples from each one, resulting in more triples compared to LLM-based pipelines. Second, the LLM-based pipelines excel at extracting gold triples, achieving significantly higher recall than tool-based pipelines (88.7\% vs. 57.84\%). Finally, in cross-lingual scenarios, LLM-based pipelines extract fewer triples from Chinese documents than from English documents  (10.29 vs. 3.58) and have a lower recall for Chinese triples (56.60\% vs. 88.51\%). This discrepancy underscores why en$\rightarrow$zh editing generally yields better results.

Furthermore, we conduct error analysis by selecting triples extracted from 100 documents across various extraction approaches. This analysis involves manual inspection, during which we categorize the errors into several types: (1) Ambiguous Triples (AT) refer to cases where the meaning of the triple is unclear due to confusing concepts or insufficient contextual information; (2) Incorrect Subject (IS), where the triple is meaningful, but the subject in the extracted triples differs from that in the gold triples; (3) Incorrect Relation (IR), where the extracted triples provide the correct subject, yet the relation does not align with that in the gold triples. We present cases of different types of errors in Table \ref{tab:triple_case}.

Back to Table \ref{tab:triplet_analysis}, the Tool-based pipeline reveals a significant proportion of Incorrect Subjects (42.74\%) and Ambiguous Triples (33.87\%). This suggests that the triples extracted through the Tool-based pipeline often fail to capture the target entities and may introduce numerous ambiguous instances. To reduce ambiguous triples, it can be helpful to train a classifier to filter them out. The challenge of Incorrect Subjects might be addressed by integrating prior knowledge, such as pre-defining a subject and filtering out unrelated triples.

In contrast, the LLM-based pipeline shows a low proportion of Ambiguous Triples (12.88\%), indicating generally higher quality in the extracted triples. However, it has a relatively high rate of Incorrect Relations, suggesting that LLMs often emphasize alternative relations associated with the correct subjects. We believe that incorporating prior relations could further enhance this approach.

For the LLM-based (zh), our error analysis reveals that the triples extracted from Chinese data contain very few ambiguous triples, with error patterns similar to those in LLM-based (en). Therefore, we believe that the suboptimal performance of the LLM-based (zh)  results from the limited number of triples extracted, which decreases the likelihood of capturing ground truth triples. To address this issue, encouraging LLMs to extract more triples from Chinese documents can lead to a possible solution.

\subsubsection{Ablation Study for Extract-then-Edit Pipeline}
\label{ablation}

\begin{table}[!h]
\small
\caption{Ablation Study for Extract-then-Edit Pipeline. \#Tri denotes the average number of triples extracted from each document.}
\renewcommand{\arraystretch}{1.2}
\centering
\begin{tabularx}{0.55\textwidth}{lcccc}
\toprule
& \textbf{\#Tri} & \textbf{Recall} & \textbf{Precision} & \textbf{F1} \\
\midrule
Tool-based & 16.50 & 57.84\% & 3.75\% & 7.04\%   \\
Tool-based w/o TV & 22.19 & 64.10\% & 3.59\% & 6.80\%   \\
Tool-based w/o SD & 39.01 & 0.30\% & 0.02\% & 0.04\%   \\
\midrule
LLM-based & 10.26 & 88.79\% & 14.71\% & 25.24\% \\
LLM-based w/o TV & 10.71 & 89.84\% & 14.48\% & 24.94\%   \\
LLM-based w/o SD & 13.71 & 88.60\% & 12.85\% & 22.44\%   \\
\bottomrule
\end{tabularx}
\label{tab:ablation_analysis}
\end{table}

In this section, we discuss the roles of various steps in our Extract-then-Edit pipeline. Specifically, we analyze the impact of two key steps: (1) Triplet Validation (TV), which uses a NER tool to eliminate low-quality triples that lack entities, and (2) Semantic Deduplication (SD), which removes duplicates by comparing semantic representations generated. We evaluate the quality of the triples extracted from the pipeline without these two steps.

The results of the ablation study for the pipeline are presented in Table \ref{tab:ablation_analysis}. For the Tool-based pipeline, the Semantic Deduplication step is particularly significant, as its removal increases the average number of triples extracted from each document from 16.5 to 39.01, while the F1 score drops from 7.04 to 0.04. In the LLM-based pipeline, the removal of both steps also leads to an increase in the number of extracted triples. Overall, both steps play a critical role in enhancing the quality of the triples and reducing the cost of knowledge editing.

\begin{figure}[!th]
    \centering
    \begin{subfigure}[b]{0.497\textwidth}
        \centering
    \includegraphics[width=\textwidth, height=0.246\textheight]{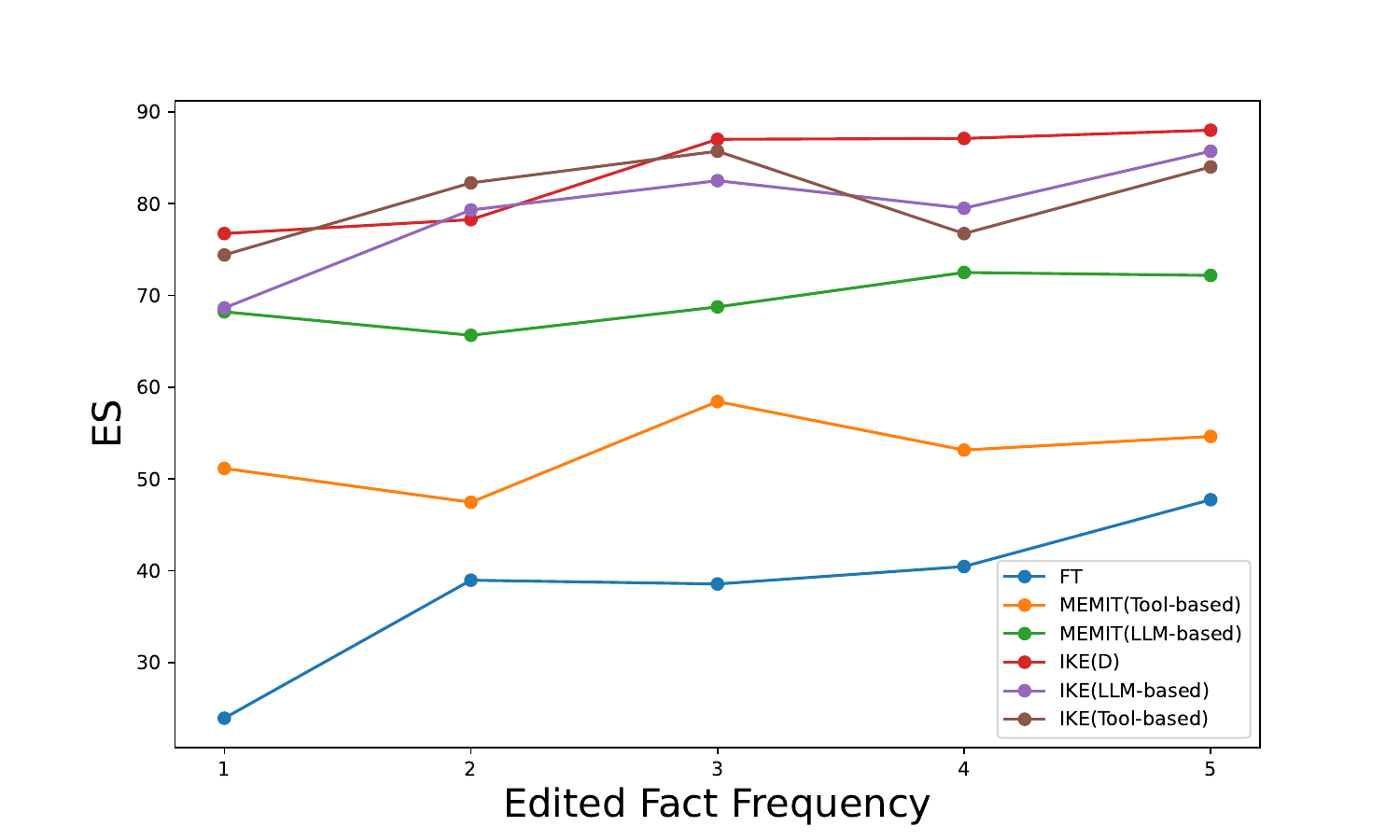}
    \end{subfigure}
    \hfill
    \begin{subfigure}[b]{0.498\textwidth}
        \centering
    \includegraphics[width=\textwidth, height=0.225\textheight]{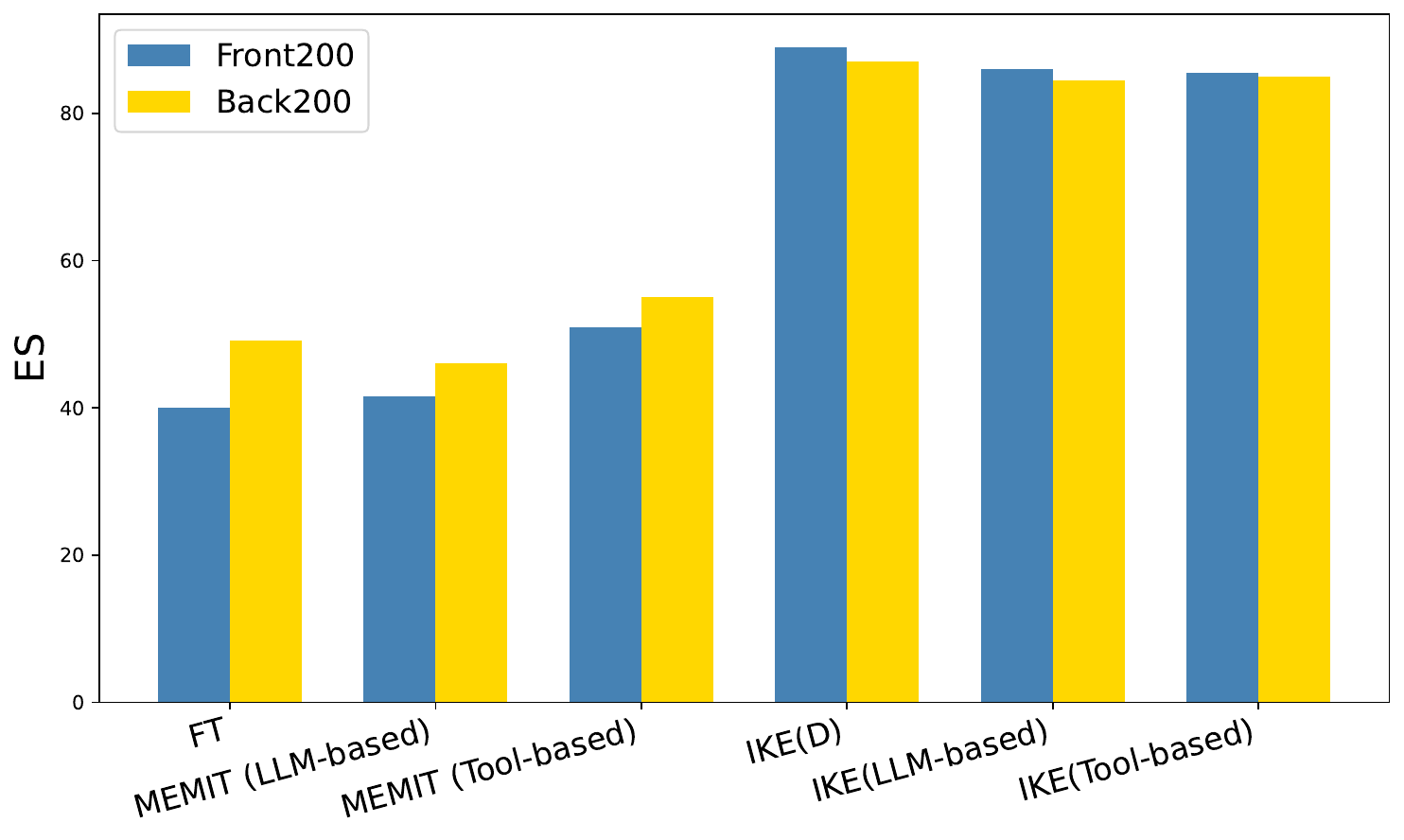}
    \end{subfigure}
    
    \caption{Left: Performance variation of knowledge editing methods with different edited fact frequency. Right: Performance of knowledge editing methods on the Front200 and Back200. Note that these two subsets exhibit significant differences in the altered target positions.}
    \label{fig:freq_pos_figs}
\end{figure}

\subsubsection{Impact of Edited Fact Frequency}
\label{sec:fact_count}
For edited facts that may be mentioned multiple times, we aim to explore how the frequency of these edited facts affects knowledge editing. As a previous study \citep{elsahar-etal-2018-rex} shows that a triplet often appears when the subject and object co-occur in text, we use the occurrence of sentences containing both these components to represent the frequency of the edit fact.  Based on Figure \ref{fig:freq_pos_figs}, we observe that the performance of all knowledge editing methods generally improves with the increased frequency of edited facts. For parameter-tuning methods like FT and MEMIT, higher frequencies of edited facts provide multiple opportunities for the LLMs to encounter and edit, aligning with findings that higher frequency knowledge is easier to learn \citep{kandpal2023longtail, carlini2023quantifying,DBLP:conf/icml/GhosalHR24}. For IKE methods, the repeated occurrence of edited facts in the context allows LLMs to focus on them more easily, resulting in better editing performance.

\subsubsection{The Impact of Altered Target Position}
\label{sec:fact_position}
The position of an altered target within a document can vary significantly (see Figure \ref{Fig_overview}), which may also affect knowledge editing. Due to the left-to-right autoregressive modeling of LLMs, later tokens can reference earlier ones. However, establishing the reverse connection poses a challenge, known as the Reversal Curse. \citep{berglund2023reversal}. For example, if the edited fact is ``\textit{What is the twin city of Manila? It is Lyon}'', the edited LLM won't automatically answer ``\textit{The twin city of Lyon is \underline{\quad}}''. This is because while learning ``\textit{Manila}'', the model cannot attend to the later-occurring ``\textit{Lyon}''. Consequently, ``\textit{Lyon}'' cannot trigger the memory of ``\textit{Manila}'' during testing. This issue also arises when extracting triples using the Extract-then-Edit pipelines.

To explore the impact of altered target position on knowledge editing, we sort documents by the position of the altered target and create two subsets: Front200, representing the 200 documents where the altered target appears earliest, and Back200 containing the latest 200 documents. We evaluate their Edit Success performance using InternLM-7B as shown in Table \ref{tab:triplet_analysis}. We find that FT and MEMIT achieve significantly higher ES scores on Back200 than on Front200. Parameter-tuning methods like FT and MEMIT depend on token order, so the later the altered target appears, the more tokens it can connect with. In contrast, IKE treats fact triples as context, allowing the LLM to consistently attend to edited knowledge when predicting, making it less sensitive to the positions of altered targets.

\section{Implications}
We investigate the under-explored document-based knowledge editing scenario by introducing a benchmark and highlighting existing issues and factors affecting performance. Our findings not only enrich the comprehension of document-based knowledge editing but also provide critical insights for future research.

First, we propose using documents for knowledge editing, which offers a more practical and cost-effective solution. 
Compared to earlier research that heavily depended on manually extracted triples (as shown in Table \ref{tab:bench_mark_comparison}), document-based knowledge editing enables the direct use of easily accessible raw documents to update the knowledge of LLMs. Thus, it is more time-efficient and adaptable to the swift pace of information updates.

Second, our work identifies issues with current document-based knowledge editing and highlights factors that influence effectiveness. We find that document-based knowledge editing is more challenging than triplet-based and performs poorly in terms of reasoning and cross-lingual transfer. Additionally, the frequency and position of factual edits in documents significantly impact editing effectiveness, consistent with the findings of research on LLM memory mechanisms \citep{kandpal2023longtail,berglund2023reversal}.

\section{Conclusion}
In this paper, we explore a more universal scenario for knowledge editing and propose \textit{DocTER}, a novel benchmark tailored for document-based knowledge editing on LLMs. This benchmark comprises a corresponding dataset for editing documents and diverse evaluation perspectives. Particularly, we assess the updated LLM in utilizing altered knowledge for reasoning and cross-lingual transfer abilities. 

Our experimental results indicate that editing with documents presents greater challenges than using triples, while also emphasizing the critical role of extracted triple quality in performance. Among the editing approaches, those leveraging external memory demonstrate superior performance but still encounter difficulties in reasoning and cross-lingual transfer. Furthermore, we examine how the quality of triplet extraction, the frequency of edited facts, and the position of altered targets affect knowledge editing performance. In our future work, we aim to delve deeper into document-based knowledge editing and explore ways to further improve reasoning and enhance cross-lingual transfer capabilities.

\section*{Acknowledgments}
The project was supported by National Natural Science Foundation of China (No. 62276219),  Natural Science Foundation of Fujian Province of China (No. 2024J011001), and the Public Technology Service Platform Project of Xiamen (No.3502Z20231043). We also thank the reviewers for their insightful comments.
\bibliographystyle{elsarticle-harv} 
\bibliography{custom}

\begin{thebibliography}{59}
\expandafter\ifx\csname natexlab\endcsname\relax\def\natexlab#1{#1}\fi
\providecommand{\url}[1]{\texttt{#1}}
\providecommand{\href}[2]{#2}
\providecommand{\path}[1]{#1}
\providecommand{\DOIprefix}{doi:}
\providecommand{\ArXivprefix}{arXiv:}
\providecommand{\URLprefix}{URL: }
\providecommand{\Pubmedprefix}{pmid:}
\providecommand{\doi}[1]{\href{http://dx.doi.org/#1}{\path{#1}}}
\providecommand{\Pubmed}[1]{\href{pmid:#1}{\path{#1}}}
\providecommand{\bibinfo}[2]{#2}
\ifx\xfnm\relax \def\xfnm[#1]{\unskip,\space#1}\fi
\bibitem[{Berglund et~al.(2024)Berglund, Tong, Kaufmann, Balesni, Stickland, Korbak and Evans}]{berglund2023reversal}
\bibinfo{author}{Berglund, L.}, \bibinfo{author}{Tong, M.}, \bibinfo{author}{Kaufmann, M.}, \bibinfo{author}{Balesni, M.}, \bibinfo{author}{Stickland, A.C.}, \bibinfo{author}{Korbak, T.}, \bibinfo{author}{Evans, O.}, \bibinfo{year}{2024}.
\newblock \bibinfo{title}{The reversal curse: Llms trained on "a is b" fail to learn "b is a"}, in: \bibinfo{booktitle}{The Twelfth International Conference on Learning Representations, {ICLR} 2024, Vienna, Austria, May 7-11, 2024}, \bibinfo{publisher}{OpenReview.net}.
\newblock \URLprefix \url{https://openreview.net/forum?id=GPKTIktA0k}.
\bibitem[{Cao et~al.(2024)Cao, Lin, Han and Sun}]{CaoLHS24}
\bibinfo{author}{Cao, B.}, \bibinfo{author}{Lin, H.}, \bibinfo{author}{Han, X.}, \bibinfo{author}{Sun, L.}, \bibinfo{year}{2024}.
\newblock \bibinfo{title}{The life cycle of knowledge in big language models: {A} survey}.
\newblock \bibinfo{journal}{Mach. Intell. Res.} \bibinfo{volume}{21}, \bibinfo{pages}{217--238}.
\newblock \URLprefix \url{https://doi.org/10.1007/s11633-023-1416-x}, \DOIprefix\doi{10.1007/S11633-023-1416-X}.
\bibitem[{Cao et~al.(2021)Cao, Aziz and Titov}]{de2021editing}
\bibinfo{author}{Cao, N.D.}, \bibinfo{author}{Aziz, W.}, \bibinfo{author}{Titov, I.}, \bibinfo{year}{2021}.
\newblock \bibinfo{title}{Editing factual knowledge in language models}, in: \bibinfo{editor}{Moens, M.}, \bibinfo{editor}{Huang, X.}, \bibinfo{editor}{Specia, L.}, \bibinfo{editor}{Yih, S.W.} (Eds.), \bibinfo{booktitle}{Proceedings of the 2021 Conference on Empirical Methods in Natural Language Processing, {EMNLP} 2021, Virtual Event / Punta Cana, Dominican Republic, 7-11 November, 2021}, \bibinfo{publisher}{Association for Computational Linguistics}. pp. \bibinfo{pages}{6491--6506}.
\newblock \URLprefix \url{https://doi.org/10.18653/v1/2021.emnlp-main.522}, \DOIprefix\doi{10.18653/V1/2021.EMNLP-MAIN.522}.
\bibitem[{Carlini et~al.(2023)Carlini, Ippolito, Jagielski, Lee, Tram{\`{e}}r and Zhang}]{carlini2023quantifying}
\bibinfo{author}{Carlini, N.}, \bibinfo{author}{Ippolito, D.}, \bibinfo{author}{Jagielski, M.}, \bibinfo{author}{Lee, K.}, \bibinfo{author}{Tram{\`{e}}r, F.}, \bibinfo{author}{Zhang, C.}, \bibinfo{year}{2023}.
\newblock \bibinfo{title}{Quantifying memorization across neural language models}, in: \bibinfo{booktitle}{The Eleventh International Conference on Learning Representations, {ICLR} 2023, Kigali, Rwanda, May 1-5, 2023}, \bibinfo{publisher}{OpenReview.net}.
\newblock \URLprefix \url{https://openreview.net/forum?id=TatRHT\_1cK}.
\bibitem[{Chang et~al.(2024)Chang, Park, Ye, Yang, Seo, Chang and Seo}]{DBLP:journals/corr/abs-2406-11813}
\bibinfo{author}{Chang, H.}, \bibinfo{author}{Park, J.}, \bibinfo{author}{Ye, S.}, \bibinfo{author}{Yang, S.}, \bibinfo{author}{Seo, Y.}, \bibinfo{author}{Chang, D.}, \bibinfo{author}{Seo, M.}, \bibinfo{year}{2024}.
\newblock \bibinfo{title}{How do large language models acquire factual knowledge during pretraining?}
\newblock \bibinfo{journal}{CoRR} \bibinfo{volume}{abs/2406.11813}.
\newblock \URLprefix \url{https://doi.org/10.48550/arXiv.2406.11813}, \DOIprefix\doi{10.48550/ARXIV.2406.11813}, \href{http://arxiv.org/abs/2406.11813}{{\tt arXiv:2406.11813}}.
\bibitem[{Chen et~al.(2024)Chen, Huang, Li, Chen, Lai, Xu, Gu, Gu, Yao, Xiao, Yan, Wang, Torr, Song and Shu}]{DBLP:edit_inject_harm}
\bibinfo{author}{Chen, C.}, \bibinfo{author}{Huang, B.}, \bibinfo{author}{Li, Z.}, \bibinfo{author}{Chen, Z.}, \bibinfo{author}{Lai, S.}, \bibinfo{author}{Xu, X.}, \bibinfo{author}{Gu, J.}, \bibinfo{author}{Gu, J.}, \bibinfo{author}{Yao, H.}, \bibinfo{author}{Xiao, C.}, \bibinfo{author}{Yan, X.}, \bibinfo{author}{Wang, W.Y.}, \bibinfo{author}{Torr, P.}, \bibinfo{author}{Song, D.}, \bibinfo{author}{Shu, K.}, \bibinfo{year}{2024}.
\newblock \bibinfo{title}{Can editing llms inject harm?}
\newblock \bibinfo{journal}{CoRR} \bibinfo{volume}{abs/2407.20224}.
\newblock \URLprefix \url{https://doi.org/10.48550/arXiv.2407.20224}, \DOIprefix\doi{10.48550/ARXIV.2407.20224}, \href{http://arxiv.org/abs/2407.20224}{{\tt arXiv:2407.20224}}.
\bibitem[{Dai et~al.(2022)Dai, Dong, Hao, Sui, Chang and Wei}]{dai2022knowledge}
\bibinfo{author}{Dai, D.}, \bibinfo{author}{Dong, L.}, \bibinfo{author}{Hao, Y.}, \bibinfo{author}{Sui, Z.}, \bibinfo{author}{Chang, B.}, \bibinfo{author}{Wei, F.}, \bibinfo{year}{2022}.
\newblock \bibinfo{title}{Knowledge neurons in pretrained transformers}, in: \bibinfo{editor}{Muresan, S.}, \bibinfo{editor}{Nakov, P.}, \bibinfo{editor}{Villavicencio, A.} (Eds.), \bibinfo{booktitle}{Proceedings of the 60th Annual Meeting of the Association for Computational Linguistics (Volume 1: Long Papers), {ACL} 2022, Dublin, Ireland, May 22-27, 2022}, \bibinfo{publisher}{Association for Computational Linguistics}. pp. \bibinfo{pages}{8493--8502}.
\newblock \URLprefix \url{https://doi.org/10.18653/v1/2022.acl-long.581}, \DOIprefix\doi{10.18653/V1/2022.ACL-LONG.581}.
\bibitem[{Dong et~al.(2022)Dong, Dai, Song, Xu, Sui and Li}]{dong2022calibrating}
\bibinfo{author}{Dong, Q.}, \bibinfo{author}{Dai, D.}, \bibinfo{author}{Song, Y.}, \bibinfo{author}{Xu, J.}, \bibinfo{author}{Sui, Z.}, \bibinfo{author}{Li, L.}, \bibinfo{year}{2022}.
\newblock \bibinfo{title}{Calibrating factual knowledge in pretrained language models}, in: \bibinfo{editor}{Goldberg, Y.}, \bibinfo{editor}{Kozareva, Z.}, \bibinfo{editor}{Zhang, Y.} (Eds.), \bibinfo{booktitle}{Findings of the Association for Computational Linguistics: {EMNLP} 2022, Abu Dhabi, United Arab Emirates, December 7-11, 2022}, \bibinfo{publisher}{Association for Computational Linguistics}. pp. \bibinfo{pages}{5937--5947}.
\newblock \URLprefix \url{https://doi.org/10.18653/v1/2022.findings-emnlp.438}, \DOIprefix\doi{10.18653/V1/2022.FINDINGS-EMNLP.438}.
\bibitem[{Dubey and et~al.(2024)}]{DBLP:journals/corr/abs-2407-21783}
\bibinfo{author}{Dubey, A.}, \bibinfo{author}{et~al.}, \bibinfo{year}{2024}.
\newblock \bibinfo{title}{The llama 3 herd of models}.
\newblock \bibinfo{journal}{CoRR} \bibinfo{volume}{abs/2407.21783}.
\newblock \URLprefix \url{https://doi.org/10.48550/arXiv.2407.21783}, \DOIprefix\doi{10.48550/ARXIV.2407.21783}, \href{http://arxiv.org/abs/2407.21783}{{\tt arXiv:2407.21783}}.
\bibitem[{ElSahar et~al.(2018)ElSahar, Vougiouklis, Remaci, Gravier, Hare, Laforest and Simperl}]{elsahar-etal-2018-rex}
\bibinfo{author}{ElSahar, H.}, \bibinfo{author}{Vougiouklis, P.}, \bibinfo{author}{Remaci, A.}, \bibinfo{author}{Gravier, C.}, \bibinfo{author}{Hare, J.S.}, \bibinfo{author}{Laforest, F.}, \bibinfo{author}{Simperl, E.}, \bibinfo{year}{2018}.
\newblock \bibinfo{title}{T-rex: {A} large scale alignment of natural language with knowledge base triples}, in: \bibinfo{editor}{Calzolari, N.}, \bibinfo{editor}{Choukri, K.}, \bibinfo{editor}{Cieri, C.}, \bibinfo{editor}{Declerck, T.}, \bibinfo{editor}{Goggi, S.}, \bibinfo{editor}{Hasida, K.}, \bibinfo{editor}{Isahara, H.}, \bibinfo{editor}{Maegaard, B.}, \bibinfo{editor}{Mariani, J.}, \bibinfo{editor}{Mazo, H.}, \bibinfo{editor}{Moreno, A.}, \bibinfo{editor}{Odijk, J.}, \bibinfo{editor}{Piperidis, S.}, \bibinfo{editor}{Tokunaga, T.} (Eds.), \bibinfo{booktitle}{Proceedings of the Eleventh International Conference on Language Resources and Evaluation, {LREC} 2018, Miyazaki, Japan, May 7-12, 2018}, \bibinfo{publisher}{European Language Resources Association {(ELRA)}}.
\newblock \URLprefix \url{http://www.lrec-conf.org/proceedings/lrec2018/summaries/632.html}.
\bibitem[{Geva et~al.(2021)Geva, Schuster, Berant and Levy}]{geva2021transformer}
\bibinfo{author}{Geva, M.}, \bibinfo{author}{Schuster, R.}, \bibinfo{author}{Berant, J.}, \bibinfo{author}{Levy, O.}, \bibinfo{year}{2021}.
\newblock \bibinfo{title}{Transformer feed-forward layers are key-value memories}, in: \bibinfo{editor}{Moens, M.}, \bibinfo{editor}{Huang, X.}, \bibinfo{editor}{Specia, L.}, \bibinfo{editor}{Yih, S.W.} (Eds.), \bibinfo{booktitle}{Proceedings of the 2021 Conference on Empirical Methods in Natural Language Processing, {EMNLP} 2021, Virtual Event / Punta Cana, Dominican Republic, 7-11 November, 2021}, \bibinfo{publisher}{Association for Computational Linguistics}. pp. \bibinfo{pages}{5484--5495}.
\newblock \URLprefix \url{https://doi.org/10.18653/v1/2021.emnlp-main.446}, \DOIprefix\doi{10.18653/V1/2021.EMNLP-MAIN.446}.
\bibitem[{Ghosal et~al.(2024)Ghosal, Hashimoto and Raghunathan}]{DBLP:conf/icml/GhosalHR24}
\bibinfo{author}{Ghosal, G.R.}, \bibinfo{author}{Hashimoto, T.}, \bibinfo{author}{Raghunathan, A.}, \bibinfo{year}{2024}.
\newblock \bibinfo{title}{Understanding finetuning for factual knowledge extraction}, in: \bibinfo{booktitle}{Forty-first International Conference on Machine Learning, {ICML} 2024, Vienna, Austria, July 21-27, 2024}, \bibinfo{publisher}{OpenReview.net}.
\newblock \URLprefix \url{https://openreview.net/forum?id=cPsn9AcOYh}.
\bibitem[{Guan et~al.(2024)Guan, Liu, Lin, Lu, He, Han and Sun}]{GuanLL0HH024}
\bibinfo{author}{Guan, X.}, \bibinfo{author}{Liu, Y.}, \bibinfo{author}{Lin, H.}, \bibinfo{author}{Lu, Y.}, \bibinfo{author}{He, B.}, \bibinfo{author}{Han, X.}, \bibinfo{author}{Sun, L.}, \bibinfo{year}{2024}.
\newblock \bibinfo{title}{Mitigating large language model hallucinations via autonomous knowledge graph-based retrofitting}, in: \bibinfo{editor}{Wooldridge, M.J.}, \bibinfo{editor}{Dy, J.G.}, \bibinfo{editor}{Natarajan, S.} (Eds.), \bibinfo{booktitle}{Thirty-Eighth {AAAI} Conference on Artificial Intelligence, {AAAI} 2024, Thirty-Sixth Conference on Innovative Applications of Artificial Intelligence, {IAAI} 2024, Fourteenth Symposium on Educational Advances in Artificial Intelligence, {EAAI} 2014, February 20-27, 2024, Vancouver, Canada}, \bibinfo{publisher}{{AAAI} Press}. pp. \bibinfo{pages}{18126--18134}.
\newblock \URLprefix \url{https://doi.org/10.1609/aaai.v38i16.29770}, \DOIprefix\doi{10.1609/AAAI.V38I16.29770}.
\bibitem[{Han et~al.(2025)Han, Wang, Shomer, Guo, Ding, Lei, Halappanavar, Rossi, Mukherjee, Tang, He, Hua, Long, Zhao, Shah, Javari, Xia and Tang}]{DBLP:journals/corr/abs-2501-00309}
\bibinfo{author}{Han, H.}, \bibinfo{author}{Wang, Y.}, \bibinfo{author}{Shomer, H.}, \bibinfo{author}{Guo, K.}, \bibinfo{author}{Ding, J.}, \bibinfo{author}{Lei, Y.}, \bibinfo{author}{Halappanavar, M.}, \bibinfo{author}{Rossi, R.A.}, \bibinfo{author}{Mukherjee, S.}, \bibinfo{author}{Tang, X.}, \bibinfo{author}{He, Q.}, \bibinfo{author}{Hua, Z.}, \bibinfo{author}{Long, B.}, \bibinfo{author}{Zhao, T.}, \bibinfo{author}{Shah, N.}, \bibinfo{author}{Javari, A.}, \bibinfo{author}{Xia, Y.}, \bibinfo{author}{Tang, J.}, \bibinfo{year}{2025}.
\newblock \bibinfo{title}{Retrieval-augmented generation with graphs (graphrag)}.
\newblock \bibinfo{journal}{CoRR} \bibinfo{volume}{abs/2501.00309}.
\newblock \URLprefix \url{https://doi.org/10.48550/arXiv.2501.00309}, \DOIprefix\doi{10.48550/ARXIV.2501.00309}, \href{http://arxiv.org/abs/2501.00309}{{\tt arXiv:2501.00309}}.
\bibitem[{Hartvigsen et~al.(2023)Hartvigsen, Sankaranarayanan, Palangi, Kim and Ghassemi}]{DBLP:conf/nips/HartvigsenSPKG23}
\bibinfo{author}{Hartvigsen, T.}, \bibinfo{author}{Sankaranarayanan, S.}, \bibinfo{author}{Palangi, H.}, \bibinfo{author}{Kim, Y.}, \bibinfo{author}{Ghassemi, M.}, \bibinfo{year}{2023}.
\newblock \bibinfo{title}{Aging with {GRACE:} lifelong model editing with discrete key-value adaptors}, in: \bibinfo{editor}{Oh, A.}, \bibinfo{editor}{Naumann, T.}, \bibinfo{editor}{Globerson, A.}, \bibinfo{editor}{Saenko, K.}, \bibinfo{editor}{Hardt, M.}, \bibinfo{editor}{Levine, S.} (Eds.), \bibinfo{booktitle}{Advances in Neural Information Processing Systems 36: Annual Conference on Neural Information Processing Systems 2023, NeurIPS 2023, New Orleans, LA, USA, December 10 - 16, 2023}.
\newblock \URLprefix \url{http://papers.nips.cc/paper\_files/paper/2023/hash/95b6e2ff961580e03c0a662a63a71812-Abstract-Conference.html}.
\bibitem[{Hendrycks et~al.(2021)Hendrycks, Burns, Basart, Zou, Mazeika, Song and Steinhardt}]{DBLP:conf/iclr/HendrycksBBZMSS21}
\bibinfo{author}{Hendrycks, D.}, \bibinfo{author}{Burns, C.}, \bibinfo{author}{Basart, S.}, \bibinfo{author}{Zou, A.}, \bibinfo{author}{Mazeika, M.}, \bibinfo{author}{Song, D.}, \bibinfo{author}{Steinhardt, J.}, \bibinfo{year}{2021}.
\newblock \bibinfo{title}{Measuring massive multitask language understanding}, in: \bibinfo{booktitle}{9th International Conference on Learning Representations, {ICLR} 2021, Virtual Event, Austria, May 3-7, 2021}, \bibinfo{publisher}{OpenReview.net}.
\newblock \URLprefix \url{https://openreview.net/forum?id=d7KBjmI3GmQ}.
\bibitem[{Hu et~al.(2024)Hu, Cao, Chen, Liu and Zhao}]{hu2024knowledgesuperpositionunveilingfailures}
\bibinfo{author}{Hu, C.}, \bibinfo{author}{Cao, P.}, \bibinfo{author}{Chen, Y.}, \bibinfo{author}{Liu, K.}, \bibinfo{author}{Zhao, J.}, \bibinfo{year}{2024}.
\newblock \bibinfo{title}{Knowledge in superposition: Unveiling the failures of lifelong knowledge editing for large language models}.
\newblock \URLprefix \url{https://doi.org/10.48550/arXiv.2408.07413}, \DOIprefix\doi{10.48550/ARXIV.2408.07413}, \href{http://arxiv.org/abs/2408.07413}{{\tt arXiv:2408.07413}}.
\bibitem[{Huang et~al.(2023)Huang, Shen, Zhang, Zhou, Rong and Xiong}]{DBLP:conf/iclr/HuangSZZR023}
\bibinfo{author}{Huang, Z.}, \bibinfo{author}{Shen, Y.}, \bibinfo{author}{Zhang, X.}, \bibinfo{author}{Zhou, J.}, \bibinfo{author}{Rong, W.}, \bibinfo{author}{Xiong, Z.}, \bibinfo{year}{2023}.
\newblock \bibinfo{title}{Transformer-patcher: One mistake worth one neuron}, in: \bibinfo{booktitle}{The Eleventh International Conference on Learning Representations, {ICLR} 2023, Kigali, Rwanda, May 1-5, 2023}, \bibinfo{publisher}{OpenReview.net}.
\newblock \URLprefix \url{https://openreview.net/forum?id=4oYUGeGBPm}.
\bibitem[{Ji et~al.(2023a)Ji, Lee, Frieske, Yu, Su, Xu, Ishii, Bang, Madotto and Fung}]{JiLFYSXIBMF23}
\bibinfo{author}{Ji, Z.}, \bibinfo{author}{Lee, N.}, \bibinfo{author}{Frieske, R.}, \bibinfo{author}{Yu, T.}, \bibinfo{author}{Su, D.}, \bibinfo{author}{Xu, Y.}, \bibinfo{author}{Ishii, E.}, \bibinfo{author}{Bang, Y.}, \bibinfo{author}{Madotto, A.}, \bibinfo{author}{Fung, P.}, \bibinfo{year}{2023}a.
\newblock \bibinfo{title}{Survey of hallucination in natural language generation}.
\newblock \bibinfo{journal}{{ACM} Comput. Surv.} \bibinfo{volume}{55}, \bibinfo{pages}{248:1--248:38}.
\newblock \URLprefix \url{https://doi.org/10.1145/3571730}, \DOIprefix\doi{10.1145/3571730}.
\bibitem[{Ji et~al.(2023b)Ji, Yu, Xu, Lee, Ishii and Fung}]{JiYXLIF23}
\bibinfo{author}{Ji, Z.}, \bibinfo{author}{Yu, T.}, \bibinfo{author}{Xu, Y.}, \bibinfo{author}{Lee, N.}, \bibinfo{author}{Ishii, E.}, \bibinfo{author}{Fung, P.}, \bibinfo{year}{2023}b.
\newblock \bibinfo{title}{Towards mitigating {LLM} hallucination via self reflection}, in: \bibinfo{editor}{Bouamor, H.}, \bibinfo{editor}{Pino, J.}, \bibinfo{editor}{Bali, K.} (Eds.), \bibinfo{booktitle}{Findings of the Association for Computational Linguistics: {EMNLP} 2023, Singapore, December 6-10, 2023}, \bibinfo{publisher}{Association for Computational Linguistics}. pp. \bibinfo{pages}{1827--1843}.
\newblock \URLprefix \url{https://doi.org/10.18653/v1/2023.findings-emnlp.123}, \DOIprefix\doi{10.18653/V1/2023.FINDINGS-EMNLP.123}.
\bibitem[{Jiang et~al.(2020)Jiang, Xu, Araki and Neubig}]{jiang2020can}
\bibinfo{author}{Jiang, Z.}, \bibinfo{author}{Xu, F.F.}, \bibinfo{author}{Araki, J.}, \bibinfo{author}{Neubig, G.}, \bibinfo{year}{2020}.
\newblock \bibinfo{title}{How can we know what language models know}.
\newblock \bibinfo{journal}{Trans. Assoc. Comput. Linguistics} \bibinfo{volume}{8}, \bibinfo{pages}{423--438}.
\newblock \URLprefix \url{https://doi.org/10.1162/tacl_a_00324}, \DOIprefix\doi{10.1162/TACL_A_00324}.
\bibitem[{Kandpal et~al.(2023)Kandpal, Deng, Roberts, Wallace and Raffel}]{kandpal2023longtail}
\bibinfo{author}{Kandpal, N.}, \bibinfo{author}{Deng, H.}, \bibinfo{author}{Roberts, A.}, \bibinfo{author}{Wallace, E.}, \bibinfo{author}{Raffel, C.}, \bibinfo{year}{2023}.
\newblock \bibinfo{title}{Large language models struggle to learn long-tail knowledge}, in: \bibinfo{editor}{Krause, A.}, \bibinfo{editor}{Brunskill, E.}, \bibinfo{editor}{Cho, K.}, \bibinfo{editor}{Engelhardt, B.}, \bibinfo{editor}{Sabato, S.}, \bibinfo{editor}{Scarlett, J.} (Eds.), \bibinfo{booktitle}{International Conference on Machine Learning, {ICML} 2023, 23-29 July 2023, Honolulu, Hawaii, {USA}}, \bibinfo{publisher}{{PMLR}}. pp. \bibinfo{pages}{15696--15707}.
\newblock \URLprefix \url{https://proceedings.mlr.press/v202/kandpal23a.html}.
\bibitem[{Kolluru et~al.(2022)Kolluru, Mohammed, Mittal, Chakrabarti and Mausam}]{kolluru-etal-2022-alignment}
\bibinfo{author}{Kolluru, K.}, \bibinfo{author}{Mohammed, M.}, \bibinfo{author}{Mittal, S.}, \bibinfo{author}{Chakrabarti, S.}, \bibinfo{author}{Mausam}, \bibinfo{year}{2022}.
\newblock \bibinfo{title}{Alignment-augmented consistent translation for multilingual open information extraction}, in: \bibinfo{editor}{Muresan, S.}, \bibinfo{editor}{Nakov, P.}, \bibinfo{editor}{Villavicencio, A.} (Eds.), \bibinfo{booktitle}{Proceedings of the 60th Annual Meeting of the Association for Computational Linguistics (Volume 1: Long Papers), {ACL} 2022, Dublin, Ireland, May 22-27, 2022}, \bibinfo{publisher}{Association for Computational Linguistics}. pp. \bibinfo{pages}{2502--2517}.
\newblock \URLprefix \url{https://doi.org/10.18653/v1/2022.acl-long.179}, \DOIprefix\doi{10.18653/V1/2022.ACL-LONG.179}.
\bibitem[{Levy et~al.(2017)Levy, Seo, Choi and Zettlemoyer}]{levy-etal-2017-zero}
\bibinfo{author}{Levy, O.}, \bibinfo{author}{Seo, M.}, \bibinfo{author}{Choi, E.}, \bibinfo{author}{Zettlemoyer, L.}, \bibinfo{year}{2017}.
\newblock \bibinfo{title}{Zero-shot relation extraction via reading comprehension}, in: \bibinfo{editor}{Levy, R.}, \bibinfo{editor}{Specia, L.} (Eds.), \bibinfo{booktitle}{Proceedings of the 21st Conference on Computational Natural Language Learning (CoNLL 2017), Vancouver, Canada, August 3-4, 2017}, \bibinfo{publisher}{Association for Computational Linguistics}. pp. \bibinfo{pages}{333--342}.
\newblock \URLprefix \url{https://doi.org/10.18653/v1/K17-1034}, \DOIprefix\doi{10.18653/V1/K17-1034}.
\bibitem[{Lewis et~al.(2020)Lewis, Perez, Piktus, Petroni, Karpukhin, Goyal, K\"{u}ttler, Lewis, Yih, Rockt\"{a}schel, Riedel and Kiela}]{NEURIPS2020_6b493230}
\bibinfo{author}{Lewis, P.}, \bibinfo{author}{Perez, E.}, \bibinfo{author}{Piktus, A.}, \bibinfo{author}{Petroni, F.}, \bibinfo{author}{Karpukhin, V.}, \bibinfo{author}{Goyal, N.}, \bibinfo{author}{K\"{u}ttler, H.}, \bibinfo{author}{Lewis, M.}, \bibinfo{author}{Yih, W.t.}, \bibinfo{author}{Rockt\"{a}schel, T.}, \bibinfo{author}{Riedel, S.}, \bibinfo{author}{Kiela, D.}, \bibinfo{year}{2020}.
\newblock \bibinfo{title}{Retrieval-augmented generation for knowledge-intensive nlp tasks}, in: \bibinfo{editor}{Larochelle, H.}, \bibinfo{editor}{Ranzato, M.}, \bibinfo{editor}{Hadsell, R.}, \bibinfo{editor}{Balcan, M.}, \bibinfo{editor}{Lin, H.} (Eds.), \bibinfo{booktitle}{Advances in Neural Information Processing Systems}, \bibinfo{publisher}{Curran Associates, Inc.}. pp. \bibinfo{pages}{9459--9474}.
\newblock \URLprefix \url{https://proceedings.neurips.cc/paper_files/paper/2020/file/6b493230205f780e1bc26945df7481e5-Paper.pdf}.
\bibitem[{Li et~al.(2024a)Li, Li, Song, Yang, Ma and Yu}]{DBLP:conf/aaai/Li0SYMY24}
\bibinfo{author}{Li, X.}, \bibinfo{author}{Li, S.}, \bibinfo{author}{Song, S.}, \bibinfo{author}{Yang, J.}, \bibinfo{author}{Ma, J.}, \bibinfo{author}{Yu, J.}, \bibinfo{year}{2024}a.
\newblock \bibinfo{title}{{PMET:} precise model editing in a transformer}, in: \bibinfo{editor}{Wooldridge, M.J.}, \bibinfo{editor}{Dy, J.G.}, \bibinfo{editor}{Natarajan, S.} (Eds.), \bibinfo{booktitle}{Thirty-Eighth {AAAI} Conference on Artificial Intelligence, {AAAI} 2024, Thirty-Sixth Conference on Innovative Applications of Artificial Intelligence, {IAAI} 2024, Fourteenth Symposium on Educational Advances in Artificial Intelligence, {EAAI} 2014, February 20-27, 2024, Vancouver, Canada}, \bibinfo{publisher}{{AAAI} Press}. pp. \bibinfo{pages}{18564--18572}.
\newblock \URLprefix \url{https://doi.org/10.1609/aaai.v38i17.29818}, \DOIprefix\doi{10.1609/AAAI.V38I17.29818}.
\bibitem[{Li et~al.(2024b)Li, Lin, Lu, Xiang, Han and Sun}]{LiL0XH024}
\bibinfo{author}{Li, Z.}, \bibinfo{author}{Lin, H.}, \bibinfo{author}{Lu, Y.}, \bibinfo{author}{Xiang, H.}, \bibinfo{author}{Han, X.}, \bibinfo{author}{Sun, L.}, \bibinfo{year}{2024}b.
\newblock \bibinfo{title}{Meta-cognitive analysis: Evaluating declarative and procedural knowledge in datasets and large language models}, in: \bibinfo{editor}{Calzolari, N.}, \bibinfo{editor}{Kan, M.}, \bibinfo{editor}{Hoste, V.}, \bibinfo{editor}{Lenci, A.}, \bibinfo{editor}{Sakti, S.}, \bibinfo{editor}{Xue, N.} (Eds.), \bibinfo{booktitle}{Proceedings of the 2024 Joint International Conference on Computational Linguistics, Language Resources and Evaluation, {LREC/COLING} 2024, 20-25 May, 2024, Torino, Italy}, \bibinfo{publisher}{{ELRA} and {ICCL}}. pp. \bibinfo{pages}{11222--11228}.
\newblock \URLprefix \url{https://aclanthology.org/2024.lrec-main.980}.
\bibitem[{Liang et~al.(2025)Liang, Bo, Gui, Zhu, Zhong, Zhao, Sun, Zhang, Zhou, Chen, Zhang and Chen}]{DBLP:conf/www/LiangBGZZZSZZCZ25}
\bibinfo{author}{Liang, L.}, \bibinfo{author}{Bo, Z.}, \bibinfo{author}{Gui, Z.}, \bibinfo{author}{Zhu, Z.}, \bibinfo{author}{Zhong, L.}, \bibinfo{author}{Zhao, P.}, \bibinfo{author}{Sun, M.}, \bibinfo{author}{Zhang, Z.}, \bibinfo{author}{Zhou, J.}, \bibinfo{author}{Chen, W.}, \bibinfo{author}{Zhang, W.}, \bibinfo{author}{Chen, H.}, \bibinfo{year}{2025}.
\newblock \bibinfo{title}{{KAG:} boosting llms in professional domains via knowledge augmented generation}, in: \bibinfo{editor}{Long, G.}, \bibinfo{editor}{Blumestein, M.}, \bibinfo{editor}{Chang, Y.}, \bibinfo{editor}{Lewin{-}Eytan, L.}, \bibinfo{editor}{Huang, Z.H.}, \bibinfo{editor}{Yom{-}Tov, E.} (Eds.), \bibinfo{booktitle}{Companion Proceedings of the {ACM} on Web Conference 2025, {WWW} 2025, Sydney, NSW, Australia, 28 April 2025 - 2 May 2025}, \bibinfo{publisher}{{ACM}}. pp. \bibinfo{pages}{334--343}.
\newblock \URLprefix \url{https://doi.org/10.1145/3701716.3715240}, \DOIprefix\doi{10.1145/3701716.3715240}.
\bibitem[{Mela et~al.(2024)Mela, Gonzalez{-}Agirre, Hernando and Villegas}]{DBLP:conf/acl/MelaGHV24}
\bibinfo{author}{Mela, D.}, \bibinfo{author}{Gonzalez{-}Agirre, A.}, \bibinfo{author}{Hernando, J.}, \bibinfo{author}{Villegas, M.}, \bibinfo{year}{2024}.
\newblock \bibinfo{title}{Mass-editing memory with attention in transformers: {A} cross-lingual exploration of knowledge}, in: \bibinfo{editor}{Ku, L.}, \bibinfo{editor}{Martins, A.}, \bibinfo{editor}{Srikumar, V.} (Eds.), \bibinfo{booktitle}{Findings of the Association for Computational Linguistics, {ACL} 2024, Bangkok, Thailand and virtual meeting, August 11-16, 2024}, \bibinfo{publisher}{Association for Computational Linguistics}. pp. \bibinfo{pages}{5831--5847}.
\newblock \URLprefix \url{https://aclanthology.org/2024.findings-acl.347}.
\bibitem[{Meng et~al.(2022)Meng, Bau, Andonian and Belinkov}]{meng2022locating}
\bibinfo{author}{Meng, K.}, \bibinfo{author}{Bau, D.}, \bibinfo{author}{Andonian, A.}, \bibinfo{author}{Belinkov, Y.}, \bibinfo{year}{2022}.
\newblock \bibinfo{title}{Locating and editing factual associations in {GPT}}, in: \bibinfo{editor}{Koyejo, S.}, \bibinfo{editor}{Mohamed, S.}, \bibinfo{editor}{Agarwal, A.}, \bibinfo{editor}{Belgrave, D.}, \bibinfo{editor}{Cho, K.}, \bibinfo{editor}{Oh, A.} (Eds.), \bibinfo{booktitle}{Advances in Neural Information Processing Systems 35: Annual Conference on Neural Information Processing Systems 2022, NeurIPS 2022, New Orleans, LA, USA, November 28 - December 9, 2022}.
\newblock \URLprefix \url{http://papers.nips.cc/paper\_files/paper/2022/hash/6f1d43d5a82a37e89b0665b33bf3a182-Abstract-Conference.html}.
\bibitem[{Meng et~al.(2023)Meng, Sharma, Andonian, Belinkov and Bau}]{meng2022mass}
\bibinfo{author}{Meng, K.}, \bibinfo{author}{Sharma, A.S.}, \bibinfo{author}{Andonian, A.J.}, \bibinfo{author}{Belinkov, Y.}, \bibinfo{author}{Bau, D.}, \bibinfo{year}{2023}.
\newblock \bibinfo{title}{Mass-editing memory in a transformer}, in: \bibinfo{booktitle}{The Eleventh International Conference on Learning Representations, {ICLR} 2023, Kigali, Rwanda, May 1-5, 2023}, \bibinfo{publisher}{OpenReview.net}.
\newblock \URLprefix \url{https://openreview.net/forum?id=MkbcAHIYgyS}.
\bibitem[{Mitchell et~al.(2022a)Mitchell, Lin, Bosselut, Finn and Manning}]{mitchell2022fast}
\bibinfo{author}{Mitchell, E.}, \bibinfo{author}{Lin, C.}, \bibinfo{author}{Bosselut, A.}, \bibinfo{author}{Finn, C.}, \bibinfo{author}{Manning, C.D.}, \bibinfo{year}{2022}a.
\newblock \bibinfo{title}{Fast model editing at scale}, in: \bibinfo{booktitle}{The Tenth International Conference on Learning Representations, {ICLR} 2022, Virtual Event, April 25-29, 2022}, \bibinfo{publisher}{OpenReview.net}.
\newblock \URLprefix \url{https://openreview.net/forum?id=0DcZxeWfOPt}.
\bibitem[{Mitchell et~al.(2022b)Mitchell, Lin, Bosselut, Manning and Finn}]{pmlr-v162-mitchell22a}
\bibinfo{author}{Mitchell, E.}, \bibinfo{author}{Lin, C.}, \bibinfo{author}{Bosselut, A.}, \bibinfo{author}{Manning, C.D.}, \bibinfo{author}{Finn, C.}, \bibinfo{year}{2022}b.
\newblock \bibinfo{title}{Memory-based model editing at scale}, in: \bibinfo{booktitle}{International Conference on Machine Learning, {ICML} 2022, 17-23 July 2022, Baltimore, Maryland, {USA}}, \bibinfo{publisher}{{PMLR}}. pp. \bibinfo{pages}{15817--15831}.
\newblock \URLprefix \url{https://proceedings.mlr.press/v162/mitchell22a.html}.
\bibitem[{Onoe et~al.(2023)Onoe, Zhang, Padmanabhan, Durrett and Choi}]{onoe-etal-2023-lms}
\bibinfo{author}{Onoe, Y.}, \bibinfo{author}{Zhang, M.J.Q.}, \bibinfo{author}{Padmanabhan, S.}, \bibinfo{author}{Durrett, G.}, \bibinfo{author}{Choi, E.}, \bibinfo{year}{2023}.
\newblock \bibinfo{title}{Can lms learn new entities from descriptions? challenges in propagating injected knowledge}, in: \bibinfo{editor}{Rogers, A.}, \bibinfo{editor}{Boyd{-}Graber, J.L.}, \bibinfo{editor}{Okazaki, N.} (Eds.), \bibinfo{booktitle}{Proceedings of the 61st Annual Meeting of the Association for Computational Linguistics (Volume 1: Long Papers), {ACL} 2023, Toronto, Canada, July 9-14, 2023}, \bibinfo{publisher}{Association for Computational Linguistics}. pp. \bibinfo{pages}{5469--5485}.
\newblock \URLprefix \url{https://doi.org/10.18653/v1/2023.acl-long.300}, \DOIprefix\doi{10.18653/V1/2023.ACL-LONG.300}.
\bibitem[{OpenAI(2023)}]{openai2023gpt4}
\bibinfo{author}{OpenAI}, \bibinfo{year}{2023}.
\newblock \bibinfo{title}{{GPT-4} technical report}.
\newblock \bibinfo{journal}{CoRR} \bibinfo{volume}{abs/2303.08774}.
\newblock \URLprefix \url{https://doi.org/10.48550/arXiv.2303.08774}, \DOIprefix\doi{10.48550/ARXIV.2303.08774}, \href{http://arxiv.org/abs/2303.08774}{{\tt arXiv:2303.08774}}.
\bibitem[{Patel et~al.(2023)Patel, Li, Rasooli, Constant, Raffel and Callison{-}Burch}]{brown2020language}
\bibinfo{author}{Patel, A.}, \bibinfo{author}{Li, B.}, \bibinfo{author}{Rasooli, M.S.}, \bibinfo{author}{Constant, N.}, \bibinfo{author}{Raffel, C.}, \bibinfo{author}{Callison{-}Burch, C.}, \bibinfo{year}{2023}.
\newblock \bibinfo{title}{Bidirectional language models are also few-shot learners}, in: \bibinfo{booktitle}{The Eleventh International Conference on Learning Representations, {ICLR} 2023, Kigali, Rwanda, May 1-5, 2023}, \bibinfo{publisher}{OpenReview.net}.
\newblock \URLprefix \url{https://openreview.net/forum?id=wCFB37bzud4}.
\bibitem[{Petroni et~al.(2019)Petroni, Rockt{\"{a}}schel, Riedel, Lewis, Bakhtin, Wu and Miller}]{petroni2019language}
\bibinfo{author}{Petroni, F.}, \bibinfo{author}{Rockt{\"{a}}schel, T.}, \bibinfo{author}{Riedel, S.}, \bibinfo{author}{Lewis, P.S.H.}, \bibinfo{author}{Bakhtin, A.}, \bibinfo{author}{Wu, Y.}, \bibinfo{author}{Miller, A.H.}, \bibinfo{year}{2019}.
\newblock \bibinfo{title}{Language models as knowledge bases?}, in: \bibinfo{editor}{Inui, K.}, \bibinfo{editor}{Jiang, J.}, \bibinfo{editor}{Ng, V.}, \bibinfo{editor}{Wan, X.} (Eds.), \bibinfo{booktitle}{Proceedings of the 2019 Conference on Empirical Methods in Natural Language Processing and the 9th International Joint Conference on Natural Language Processing, {EMNLP-IJCNLP} 2019, Hong Kong, China, November 3-7, 2019}, \bibinfo{publisher}{Association for Computational Linguistics}. pp. \bibinfo{pages}{2463--2473}.
\newblock \URLprefix \url{https://doi.org/10.18653/v1/D19-1250}, \DOIprefix\doi{10.18653/V1/D19-1250}.
\bibitem[{Raffel et~al.(2020)Raffel, Shazeer, Roberts, Lee, Narang, Matena, Zhou, Li and Liu}]{DBLP:journals/jmlr/RaffelSRLNMZLL20}
\bibinfo{author}{Raffel, C.}, \bibinfo{author}{Shazeer, N.}, \bibinfo{author}{Roberts, A.}, \bibinfo{author}{Lee, K.}, \bibinfo{author}{Narang, S.}, \bibinfo{author}{Matena, M.}, \bibinfo{author}{Zhou, Y.}, \bibinfo{author}{Li, W.}, \bibinfo{author}{Liu, P.J.}, \bibinfo{year}{2020}.
\newblock \bibinfo{title}{Exploring the limits of transfer learning with a unified text-to-text transformer}.
\newblock \bibinfo{journal}{J. Mach. Learn. Res.} \bibinfo{volume}{21}, \bibinfo{pages}{140:1--140:67}.
\newblock \URLprefix \url{https://jmlr.org/papers/v21/20-074.html}.
\bibitem[{Reimers and Gurevych(2019)}]{reimers2019sentencebertsentenceembeddingsusing}
\bibinfo{author}{Reimers, N.}, \bibinfo{author}{Gurevych, I.}, \bibinfo{year}{2019}.
\newblock \bibinfo{title}{Sentence-bert: Sentence embeddings using siamese bert-networks}, in: \bibinfo{editor}{Inui, K.}, \bibinfo{editor}{Jiang, J.}, \bibinfo{editor}{Ng, V.}, \bibinfo{editor}{Wan, X.} (Eds.), \bibinfo{booktitle}{Proceedings of the 2019 Conference on Empirical Methods in Natural Language Processing and the 9th International Joint Conference on Natural Language Processing, {EMNLP-IJCNLP} 2019, Hong Kong, China, November 3-7, 2019}, \bibinfo{publisher}{Association for Computational Linguistics}. pp. \bibinfo{pages}{3980--3990}.
\newblock \URLprefix \url{https://doi.org/10.18653/v1/D19-1410}, \DOIprefix\doi{10.18653/V1/D19-1410}.
\bibitem[{Roberts et~al.(2020)Roberts, Raffel and Shazeer}]{roberts2020much}
\bibinfo{author}{Roberts, A.}, \bibinfo{author}{Raffel, C.}, \bibinfo{author}{Shazeer, N.}, \bibinfo{year}{2020}.
\newblock \bibinfo{title}{How much knowledge can you pack into the parameters of a language model?}, in: \bibinfo{editor}{Webber, B.}, \bibinfo{editor}{Cohn, T.}, \bibinfo{editor}{He, Y.}, \bibinfo{editor}{Liu, Y.} (Eds.), \bibinfo{booktitle}{Proceedings of the 2020 Conference on Empirical Methods in Natural Language Processing, {EMNLP} 2020, Online, November 16-20, 2020}, \bibinfo{publisher}{Association for Computational Linguistics}. pp. \bibinfo{pages}{5418--5426}.
\newblock \URLprefix \url{https://doi.org/10.18653/v1/2020.emnlp-main.437}, \DOIprefix\doi{10.18653/V1/2020.EMNLP-MAIN.437}.
\bibitem[{Talmor et~al.(2019)Talmor, Herzig, Lourie and Berant}]{DBLP:conf/naacl/TalmorHLB19}
\bibinfo{author}{Talmor, A.}, \bibinfo{author}{Herzig, J.}, \bibinfo{author}{Lourie, N.}, \bibinfo{author}{Berant, J.}, \bibinfo{year}{2019}.
\newblock \bibinfo{title}{Commonsenseqa: {A} question answering challenge targeting commonsense knowledge}, in: \bibinfo{editor}{Burstein, J.}, \bibinfo{editor}{Doran, C.}, \bibinfo{editor}{Solorio, T.} (Eds.), \bibinfo{booktitle}{Proceedings of the 2019 Conference of the North American Chapter of the Association for Computational Linguistics: Human Language Technologies, {NAACL-HLT} 2019, Minneapolis, MN, USA, June 2-7, 2019, Volume 1 (Long and Short Papers)}, \bibinfo{publisher}{Association for Computational Linguistics}. pp. \bibinfo{pages}{4149--4158}.
\newblock \URLprefix \url{https://doi.org/10.18653/v1/n19-1421}, \DOIprefix\doi{10.18653/V1/N19-1421}.
\bibitem[{Team(2023)}]{2023internlm}
\bibinfo{author}{Team, I.}, \bibinfo{year}{2023}.
\newblock \bibinfo{title}{Internlm: A multilingual language model with progressively enhanced capabilities}.
\newblock \bibinfo{howpublished}{\url{https://github.com/InternLM/InternLM-techreport}}.
\bibitem[{Thorne et~al.(2018)Thorne, Vlachos, Christodoulopoulos and Mittal}]{thorne-etal-2018-fever}
\bibinfo{author}{Thorne, J.}, \bibinfo{author}{Vlachos, A.}, \bibinfo{author}{Christodoulopoulos, C.}, \bibinfo{author}{Mittal, A.}, \bibinfo{year}{2018}.
\newblock \bibinfo{title}{{FEVER:} a large-scale dataset for fact extraction and verification}, in: \bibinfo{editor}{Walker, M.A.}, \bibinfo{editor}{Ji, H.}, \bibinfo{editor}{Stent, A.} (Eds.), \bibinfo{booktitle}{Proceedings of the 2018 Conference of the North American Chapter of the Association for Computational Linguistics: Human Language Technologies, {NAACL-HLT} 2018, New Orleans, Louisiana, USA, June 1-6, 2018, Volume 1 (Long Papers)}, \bibinfo{publisher}{Association for Computational Linguistics}. pp. \bibinfo{pages}{809--819}.
\newblock \URLprefix \url{https://doi.org/10.18653/v1/n18-1074}, \DOIprefix\doi{10.18653/V1/N18-1074}.
\bibitem[{Touvron et~al.(2023)Touvron, Lavril, Izacard, Martinet, Lachaux, Lacroix, Rozi{\`{e}}re, Goyal, Hambro, Azhar, Rodriguez, Joulin, Grave and Lample}]{touvron2023llama}
\bibinfo{author}{Touvron, H.}, \bibinfo{author}{Lavril, T.}, \bibinfo{author}{Izacard, G.}, \bibinfo{author}{Martinet, X.}, \bibinfo{author}{Lachaux, M.}, \bibinfo{author}{Lacroix, T.}, \bibinfo{author}{Rozi{\`{e}}re, B.}, \bibinfo{author}{Goyal, N.}, \bibinfo{author}{Hambro, E.}, \bibinfo{author}{Azhar, F.}, \bibinfo{author}{Rodriguez, A.}, \bibinfo{author}{Joulin, A.}, \bibinfo{author}{Grave, E.}, \bibinfo{author}{Lample, G.}, \bibinfo{year}{2023}.
\newblock \bibinfo{title}{Llama: Open and efficient foundation language models}.
\newblock \bibinfo{journal}{CoRR} \bibinfo{volume}{abs/2302.13971}.
\newblock \URLprefix \url{https://doi.org/10.48550/arXiv.2302.13971}, \DOIprefix\doi{10.48550/ARXIV.2302.13971}, \href{http://arxiv.org/abs/2302.13971}{{\tt arXiv:2302.13971}}.
\bibitem[{Wang et~al.(2024)Wang, Liang, Sun, Cao, Xu and Meng}]{wang2024crosslingualknowledgeediting}
\bibinfo{author}{Wang, J.}, \bibinfo{author}{Liang, Y.}, \bibinfo{author}{Sun, Z.}, \bibinfo{author}{Cao, Y.}, \bibinfo{author}{Xu, J.}, \bibinfo{author}{Meng, F.}, \bibinfo{year}{2024}.
\newblock \bibinfo{title}{Cross-lingual knowledge editing in large language models}, in: \bibinfo{editor}{Ku, L.}, \bibinfo{editor}{Martins, A.}, \bibinfo{editor}{Srikumar, V.} (Eds.), \bibinfo{booktitle}{Proceedings of the 62nd Annual Meeting of the Association for Computational Linguistics (Volume 1: Long Papers), {ACL} 2024, Bangkok, Thailand, August 11-16, 2024}, \bibinfo{publisher}{Association for Computational Linguistics}. pp. \bibinfo{pages}{11676--11686}.
\newblock \URLprefix \url{https://aclanthology.org/2024.acl-long.627}.
\bibitem[{Wang et~al.(2023)Wang, Zhang, Tian, Xi, Yao, Xu, Wang, Mao, Wang, Cheng, Liu, Ni, Zheng and Chen}]{DBLP:journals/corr/abs-2308-07269}
\bibinfo{author}{Wang, P.}, \bibinfo{author}{Zhang, N.}, \bibinfo{author}{Tian, B.}, \bibinfo{author}{Xi, Z.}, \bibinfo{author}{Yao, Y.}, \bibinfo{author}{Xu, Z.}, \bibinfo{author}{Wang, M.}, \bibinfo{author}{Mao, S.}, \bibinfo{author}{Wang, X.}, \bibinfo{author}{Cheng, S.}, \bibinfo{author}{Liu, K.}, \bibinfo{author}{Ni, Y.}, \bibinfo{author}{Zheng, G.}, \bibinfo{author}{Chen, H.}, \bibinfo{year}{2023}.
\newblock \bibinfo{title}{Easyedit: An easy-to-use knowledge editing framework for large language models}.
\newblock \bibinfo{journal}{CoRR} \bibinfo{volume}{abs/2308.07269}.
\newblock \URLprefix \url{https://doi.org/10.48550/arXiv.2308.07269}, \DOIprefix\doi{10.48550/ARXIV.2308.07269}, \href{http://arxiv.org/abs/2308.07269}{{\tt arXiv:2308.07269}}.
\bibitem[{Wei et~al.(2022)Wei, Wang, Schuurmans, Bosma, Ichter, Xia, Chi, Le and Zhou}]{DBLP:conf/nips/Wei0SBIXCLZ22}
\bibinfo{author}{Wei, J.}, \bibinfo{author}{Wang, X.}, \bibinfo{author}{Schuurmans, D.}, \bibinfo{author}{Bosma, M.}, \bibinfo{author}{Ichter, B.}, \bibinfo{author}{Xia, F.}, \bibinfo{author}{Chi, E.H.}, \bibinfo{author}{Le, Q.V.}, \bibinfo{author}{Zhou, D.}, \bibinfo{year}{2022}.
\newblock \bibinfo{title}{Chain-of-thought prompting elicits reasoning in large language models}, in: \bibinfo{editor}{Koyejo, S.}, \bibinfo{editor}{Mohamed, S.}, \bibinfo{editor}{Agarwal, A.}, \bibinfo{editor}{Belgrave, D.}, \bibinfo{editor}{Cho, K.}, \bibinfo{editor}{Oh, A.} (Eds.), \bibinfo{booktitle}{Advances in Neural Information Processing Systems 35: Annual Conference on Neural Information Processing Systems 2022, NeurIPS 2022, New Orleans, LA, USA, November 28 - December 9, 2022}.
\newblock \URLprefix \url{http://papers.nips.cc/paper\_files/paper/2022/hash/9d5609613524ecf4f15af0f7b31abca4-Abstract-Conference.html}.
\bibitem[{Wei et~al.(2024a)Wei, Zhang, Zhang, Ding, Chen, Ong, Zhang and Xiang}]{DBLP:journals/corr/abs-2406-03880}
\bibinfo{author}{Wei, J.}, \bibinfo{author}{Zhang, Y.}, \bibinfo{author}{Zhang, L.Y.}, \bibinfo{author}{Ding, M.}, \bibinfo{author}{Chen, C.}, \bibinfo{author}{Ong, K.}, \bibinfo{author}{Zhang, J.}, \bibinfo{author}{Xiang, Y.}, \bibinfo{year}{2024}a.
\newblock \bibinfo{title}{Memorization in deep learning: {A} survey}.
\newblock \bibinfo{journal}{CoRR} \bibinfo{volume}{abs/2406.03880}.
\newblock \URLprefix \url{https://doi.org/10.48550/arXiv.2406.03880}, \DOIprefix\doi{10.48550/ARXIV.2406.03880}, \href{http://arxiv.org/abs/2406.03880}{{\tt arXiv:2406.03880}}.
\bibitem[{Wei et~al.(2024b)Wei, Yu, Weng, Ma, Zhang, Zhao and Liu}]{DBLP:conf/cikm/WeiYWMZ0024}
\bibinfo{author}{Wei, Y.}, \bibinfo{author}{Yu, X.}, \bibinfo{author}{Weng, Y.}, \bibinfo{author}{Ma, H.}, \bibinfo{author}{Zhang, Y.}, \bibinfo{author}{Zhao, J.}, \bibinfo{author}{Liu, K.}, \bibinfo{year}{2024}b.
\newblock \bibinfo{title}{Does knowledge localization hold true? surprising differences between entity and relation perspectives in language models}, in: \bibinfo{editor}{Serra, E.}, \bibinfo{editor}{Spezzano, F.} (Eds.), \bibinfo{booktitle}{Proceedings of the 33rd {ACM} International Conference on Information and Knowledge Management, {CIKM} 2024, Boise, ID, USA, October 21-25, 2024}, \bibinfo{publisher}{{ACM}}. pp. \bibinfo{pages}{4118--4122}.
\newblock \URLprefix \url{https://doi.org/10.1145/3627673.3679900}, \DOIprefix\doi{10.1145/3627673.3679900}.
\bibitem[{Yan et~al.(2024)Yan, Gu, Zhu and Ling}]{yan2024correctiveretrievalaugmentedgeneration}
\bibinfo{author}{Yan, S.Q.}, \bibinfo{author}{Gu, J.C.}, \bibinfo{author}{Zhu, Y.}, \bibinfo{author}{Ling, Z.H.}, \bibinfo{year}{2024}.
\newblock \bibinfo{title}{Corrective retrieval augmented generation}.
\newblock \URLprefix \url{https://arxiv.org/abs/2401.15884}, \href{http://arxiv.org/abs/2401.15884}{{\tt arXiv:2401.15884}}.
\bibitem[{Zhang et~al.(2024a)Zhang, Yao, Tian, Wang, Deng, Wang, Xi, Mao, Zhang, Ni, Cheng, Xu, Xu, Gu, Jiang, Xie, Huang, Liang, Zhang, Zhu, Zhou and Chen}]{zhang2024comprehensive}
\bibinfo{author}{Zhang, N.}, \bibinfo{author}{Yao, Y.}, \bibinfo{author}{Tian, B.}, \bibinfo{author}{Wang, P.}, \bibinfo{author}{Deng, S.}, \bibinfo{author}{Wang, M.}, \bibinfo{author}{Xi, Z.}, \bibinfo{author}{Mao, S.}, \bibinfo{author}{Zhang, J.}, \bibinfo{author}{Ni, Y.}, \bibinfo{author}{Cheng, S.}, \bibinfo{author}{Xu, Z.}, \bibinfo{author}{Xu, X.}, \bibinfo{author}{Gu, J.}, \bibinfo{author}{Jiang, Y.}, \bibinfo{author}{Xie, P.}, \bibinfo{author}{Huang, F.}, \bibinfo{author}{Liang, L.}, \bibinfo{author}{Zhang, Z.}, \bibinfo{author}{Zhu, X.}, \bibinfo{author}{Zhou, J.}, \bibinfo{author}{Chen, H.}, \bibinfo{year}{2024}a.
\newblock \bibinfo{title}{A comprehensive study of knowledge editing for large language models}.
\newblock \bibinfo{journal}{CoRR} \bibinfo{volume}{abs/2401.01286}.
\newblock \URLprefix \url{https://doi.org/10.48550/arXiv.2401.01286}, \href{http://arxiv.org/abs/2401.01286}{{\tt arXiv:2401.01286}}.
\bibitem[{Zhang et~al.(2024b)Zhang, Wan, Zhang, Cheung, Tian, Shen and Ye}]{DBLP:conf/icml/ZhangWZC0SY24}
\bibinfo{author}{Zhang, W.}, \bibinfo{author}{Wan, C.}, \bibinfo{author}{Zhang, Y.}, \bibinfo{author}{Cheung, Y.}, \bibinfo{author}{Tian, X.}, \bibinfo{author}{Shen, X.}, \bibinfo{author}{Ye, J.}, \bibinfo{year}{2024}b.
\newblock \bibinfo{title}{Interpreting and improving large language models in arithmetic calculation}, in: \bibinfo{booktitle}{Forty-first International Conference on Machine Learning, {ICML} 2024, Vienna, Austria, July 21-27, 2024}, \bibinfo{publisher}{OpenReview.net}.
\newblock \URLprefix \url{https://openreview.net/forum?id=CfOtiepP8s}.
\bibitem[{Zhang et~al.(2023)Zhang, Li, Cui, Cai, Liu, Fu, Huang, Zhao, Zhang, Chen, Wang, Luu, Bi, Shi and Shi}]{zhang2023sirens}
\bibinfo{author}{Zhang, Y.}, \bibinfo{author}{Li, Y.}, \bibinfo{author}{Cui, L.}, \bibinfo{author}{Cai, D.}, \bibinfo{author}{Liu, L.}, \bibinfo{author}{Fu, T.}, \bibinfo{author}{Huang, X.}, \bibinfo{author}{Zhao, E.}, \bibinfo{author}{Zhang, Y.}, \bibinfo{author}{Chen, Y.}, \bibinfo{author}{Wang, L.}, \bibinfo{author}{Luu, A.T.}, \bibinfo{author}{Bi, W.}, \bibinfo{author}{Shi, F.}, \bibinfo{author}{Shi, S.}, \bibinfo{year}{2023}.
\newblock \bibinfo{title}{Siren's song in the {AI} ocean: {A} survey on hallucination in large language models}.
\newblock \bibinfo{journal}{CoRR} \bibinfo{volume}{abs/2309.01219}.
\newblock \URLprefix \url{https://doi.org/10.48550/arXiv.2309.01219}, \DOIprefix\doi{10.48550/ARXIV.2309.01219}, \href{http://arxiv.org/abs/2309.01219}{{\tt arXiv:2309.01219}}.
\bibitem[{Zhao et~al.(2023)Zhao, Zhou, Li, Tang, Wang, Hou, Min, Zhang, Zhang, Dong, Du, Yang, Chen, Chen, Jiang, Ren, Li, Tang, Liu, Liu, Nie and Wen}]{zhao2023survey}
\bibinfo{author}{Zhao, W.X.}, \bibinfo{author}{Zhou, K.}, \bibinfo{author}{Li, J.}, \bibinfo{author}{Tang, T.}, \bibinfo{author}{Wang, X.}, \bibinfo{author}{Hou, Y.}, \bibinfo{author}{Min, Y.}, \bibinfo{author}{Zhang, B.}, \bibinfo{author}{Zhang, J.}, \bibinfo{author}{Dong, Z.}, \bibinfo{author}{Du, Y.}, \bibinfo{author}{Yang, C.}, \bibinfo{author}{Chen, Y.}, \bibinfo{author}{Chen, Z.}, \bibinfo{author}{Jiang, J.}, \bibinfo{author}{Ren, R.}, \bibinfo{author}{Li, Y.}, \bibinfo{author}{Tang, X.}, \bibinfo{author}{Liu, Z.}, \bibinfo{author}{Liu, P.}, \bibinfo{author}{Nie, J.}, \bibinfo{author}{Wen, J.}, \bibinfo{year}{2023}.
\newblock \bibinfo{title}{A survey of large language models}.
\newblock \bibinfo{journal}{CoRR} \bibinfo{volume}{abs/2303.18223}.
\newblock \URLprefix \url{https://doi.org/10.48550/arXiv.2303.18223}, \DOIprefix\doi{10.48550/ARXIV.2303.18223}, \href{http://arxiv.org/abs/2303.18223}{{\tt arXiv:2303.18223}}.
\bibitem[{Zheng et~al.(2023)Zheng, Li, Dong, Fan, Wu, Xu and Chang}]{zheng-etal-2023-edit}
\bibinfo{author}{Zheng, C.}, \bibinfo{author}{Li, L.}, \bibinfo{author}{Dong, Q.}, \bibinfo{author}{Fan, Y.}, \bibinfo{author}{Wu, Z.}, \bibinfo{author}{Xu, J.}, \bibinfo{author}{Chang, B.}, \bibinfo{year}{2023}.
\newblock \bibinfo{title}{Can we edit factual knowledge by in-context learning?}, in: \bibinfo{editor}{Bouamor, H.}, \bibinfo{editor}{Pino, J.}, \bibinfo{editor}{Bali, K.} (Eds.), \bibinfo{booktitle}{Proceedings of the 2023 Conference on Empirical Methods in Natural Language Processing, {EMNLP} 2023, Singapore, December 6-10, 2023}, \bibinfo{publisher}{Association for Computational Linguistics}. pp. \bibinfo{pages}{4862--4876}.
\newblock \URLprefix \url{https://doi.org/10.18653/v1/2023.emnlp-main.296}, \DOIprefix\doi{10.18653/V1/2023.EMNLP-MAIN.296}.
\bibitem[{Zheng et~al.(2024)Zheng, Qiu, Shi and Ma}]{DBLP:journals/corr/abs-2406-06391}
\bibinfo{author}{Zheng, J.}, \bibinfo{author}{Qiu, S.}, \bibinfo{author}{Shi, C.}, \bibinfo{author}{Ma, Q.}, \bibinfo{year}{2024}.
\newblock \bibinfo{title}{Towards lifelong learning of large language models: {A} survey}.
\newblock \bibinfo{journal}{CoRR} \bibinfo{volume}{abs/2406.06391}.
\newblock \URLprefix \url{https://doi.org/10.48550/arXiv.2406.06391}, \DOIprefix\doi{10.48550/ARXIV.2406.06391}, \href{http://arxiv.org/abs/2406.06391}{{\tt arXiv:2406.06391}}.
\bibitem[{Zhong et~al.(2024)Zhong, Cui, Guo, Liang, Lu, Wang, Saied, Chen and Duan}]{DBLP:conf/naacl/ZhongCGLLWSCD24}
\bibinfo{author}{Zhong, W.}, \bibinfo{author}{Cui, R.}, \bibinfo{author}{Guo, Y.}, \bibinfo{author}{Liang, Y.}, \bibinfo{author}{Lu, S.}, \bibinfo{author}{Wang, Y.}, \bibinfo{author}{Saied, A.}, \bibinfo{author}{Chen, W.}, \bibinfo{author}{Duan, N.}, \bibinfo{year}{2024}.
\newblock \bibinfo{title}{Agieval: {A} human-centric benchmark for evaluating foundation models}, in: \bibinfo{editor}{Duh, K.}, \bibinfo{editor}{G{\'{o}}mez{-}Adorno, H.}, \bibinfo{editor}{Bethard, S.} (Eds.), \bibinfo{booktitle}{Findings of the Association for Computational Linguistics: {NAACL} 2024, Mexico City, Mexico, June 16-21, 2024}, \bibinfo{publisher}{Association for Computational Linguistics}. pp. \bibinfo{pages}{2299--2314}.
\newblock \URLprefix \url{https://doi.org/10.18653/v1/2024.findings-naacl.149}, \DOIprefix\doi{10.18653/V1/2024.FINDINGS-NAACL.149}.
\bibitem[{Zhong et~al.(2023)Zhong, Wu, Manning, Potts and Chen}]{zhong2023mquake}
\bibinfo{author}{Zhong, Z.}, \bibinfo{author}{Wu, Z.}, \bibinfo{author}{Manning, C.D.}, \bibinfo{author}{Potts, C.}, \bibinfo{author}{Chen, D.}, \bibinfo{year}{2023}.
\newblock \bibinfo{title}{Mquake: Assessing knowledge editing in language models via multi-hop questions}, in: \bibinfo{editor}{Bouamor, H.}, \bibinfo{editor}{Pino, J.}, \bibinfo{editor}{Bali, K.} (Eds.), \bibinfo{booktitle}{Proceedings of the 2023 Conference on Empirical Methods in Natural Language Processing, {EMNLP} 2023, Singapore, December 6-10, 2023}, \bibinfo{publisher}{Association for Computational Linguistics}. pp. \bibinfo{pages}{15686--15702}.
\newblock \URLprefix \url{https://doi.org/10.18653/v1/2023.emnlp-main.971}, \DOIprefix\doi{10.18653/V1/2023.EMNLP-MAIN.971}.
\bibitem[{Zhu et~al.(2020)Zhu, Rawat, Zaheer, Bhojanapalli, Li, Yu and Kumar}]{zhu2020modifying}
\bibinfo{author}{Zhu, C.}, \bibinfo{author}{Rawat, A.S.}, \bibinfo{author}{Zaheer, M.}, \bibinfo{author}{Bhojanapalli, S.}, \bibinfo{author}{Li, D.}, \bibinfo{author}{Yu, F.X.}, \bibinfo{author}{Kumar, S.}, \bibinfo{year}{2020}.
\newblock \bibinfo{title}{Modifying memories in transformer models}.
\newblock \bibinfo{journal}{CoRR} \bibinfo{volume}{abs/2012.00363}.
\newblock \URLprefix \url{https://arxiv.org/abs/2012.00363}, \href{http://arxiv.org/abs/2012.00363}{{\tt arXiv:2012.00363}}.

\end{thebibliography}
\newpage
\appendix

\label{sec:appendix}
\section{Annotation Guideline}
\label{sec:guide}
For documents or queries produced by ChatGPT, we initially prompt the model to determine if the document conveys the edited fact. If it fails, we regenerate the document, which happens for about 7\% of the cases. With three of the authors conducting manual reviews, we retain only those instances that all agree are of high quality, resulting in 74\% of the original documents and 81\% of the translated queries being preserved without revisions. The detailed Annotation Guideline is provided below.
\\
\begin{tcolorbox}[width=\linewidth, colback=white!95!black, title={Annotation Guideline}]

For generated \textbf{document}:
\begin{itemize}
\item Check if the document includes at least one mention of the \{edited fact\}. If not, discard the document.
\item Check for any knowledge conflicts related to \{edited fact\} within the document. If conflicts are found, discard the document.
\item Check if the content generated by the model is readable and fluent. If it is difficult to understand, discard the document.
\end{itemize}

For translated \textbf{Chinese queries}:
\begin{itemize}
\item Verify that the translated content matches the English query: \{en query\}. If not, discard the instance.
\item Ensure that the translated queries from the original English paraphrased queries instance all inquire about the same fact. If not, discard the instance.
\item Check if the content generated by the model is readable and fluent. If it is difficult to understand, discard the instance.
\end{itemize}

\end{tcolorbox}

\end{document}